\documentclass[10pt]{article} % For LaTeX2e
\usepackage[accepted]{tmlr}

\usepackage{amsmath,amsfonts,bm}

\def\eqref#1{equation~\ref{#1}}
\def\1{\bm{1}}

\DeclareMathAlphabet{\mathsfit}{\encodingdefault}{\sfdefault}{m}{sl}
\SetMathAlphabet{\mathsfit}{bold}{\encodingdefault}{\sfdefault}{bx}{n}

\usepackage{microtype}
\usepackage{graphicx}
\usepackage{subcaption}
\usepackage{booktabs}
\usepackage{hyperref}
\usepackage{url}
\usepackage{amsmath}
\usepackage{amssymb}
\usepackage{mathtools}
\usepackage{amsthm}
\usepackage[capitalize,noabbrev]{cleveref}
\usepackage{multirow}
\usepackage{adjustbox}
\usepackage{caption}
\usepackage{wrapfig}
\usepackage{xfp}
\usepackage{tabularx}
\usepackage[table]{xcolor}
\usepackage{algorithm}
\usepackage{algorithmic}
\usepackage[section]{placeins} % prevent floats from drifting past section boundaries

\definecolor{myrose}{HTML}{E6F0FF}
\definecolor{myblue}{HTML}{DAC3CC}

\theoremstyle{plain}

\theoremstyle{definition}

\theoremstyle{remark}

\usepackage[disable,textsize=tiny]{todonotes}

\hypersetup{
  pdftitle={POPI: Personalizing LLMs via Optimized Natural Language Preference Inference},
  pdfauthor={Yizhuo Chen, Xin Liu, Ruijie Wang, Zheng Li, Pei Chen, Changlong Yu, Qingyu Yin, Priyanka Nigam, Meng Jiang, Bing Yin},
  pdfsubject={Transactions on Machine Learning Research},
  pdfkeywords={LLM personalization, preference inference, preference optimization, natural language interface}
}

\newcommand{\tm}{POPI}
\newcommand{\ci}[1]{{\scriptsize\,$\pm$\,#1}}

\def\month{09}
\def\year{2026}
\def\openreview{\url{https://openreview.net/forum?id=KUyeGxlirZ}}

\title{POPI: Personalizing LLMs via Optimized Natural Language Preference Inference}

\author{\name Yizhuo Chen$^{1,2}$\thanks{Work done during an internship at Amazon.} \email yizhuoc@illinois.edu \\
      \name Xin Liu$^{2}$ \email xliucs@amazon.com \\
      \name Ruijie Wang$^{2}$\thanks{Corresponding author.} \email ruijiew@amazon.com \\
      \name Zheng Li$^{2}$ \email amznlz@amazon.com \\
      \name Pei Chen$^{2}$ \email peiachen@amazon.com \\
      \name Changlong Yu$^{2}$ \email changlony@amazon.com \\
      \name Qingyu Yin$^{2}$ \email qingyuyin@amazon.com \\
      \name Priyanka Nigam$^{2}$ \email priyniga@amazon.com \\
      \name Meng Jiang$^{3,2}$ \email mjiang2@nd.edu \\
      \name Bing Yin$^{2}$ \email bingyin@amazon.com \\
      \addr $^{1}$ University of Illinois Urbana-Champaign, Urbana, IL, USA \\
      $^{2}$ Amazon, USA \\
      $^{3}$ University of Notre Dame, Notre Dame, IN, USA}

\begin{document}

\maketitle

\begin{abstract}
    Large language models (LLMs) are typically aligned with population-level preferences, despite substantial variation across individual users. We introduce \tm{}, a user-level personalization framework that separates the problem into two components connected by a natural-language interface: a shared inference model that distills heterogeneous user signals into a concise preference summary, and a shared generator that conditions on this summary to produce personalized responses. Both components are trained under a unified preference-optimization objective, with reinforcement learning handling the non-differentiable inference step. This objective decomposes into generator approximation error and summary informativeness, revealing how a single loss simultaneously drives accurate generation and informative summarization. Because the interface is natural language, learned summaries can be inferred once per user and reused across different generators---including frozen, black-box commercial APIs. Across four personalization benchmarks, \tm{} generally improves personalization quality while reducing context overhead by up to an order of magnitude.
    \end{abstract}

\section{Introduction}

\begin{figure}[!ht]
    \centering
    \includegraphics[width=\textwidth]{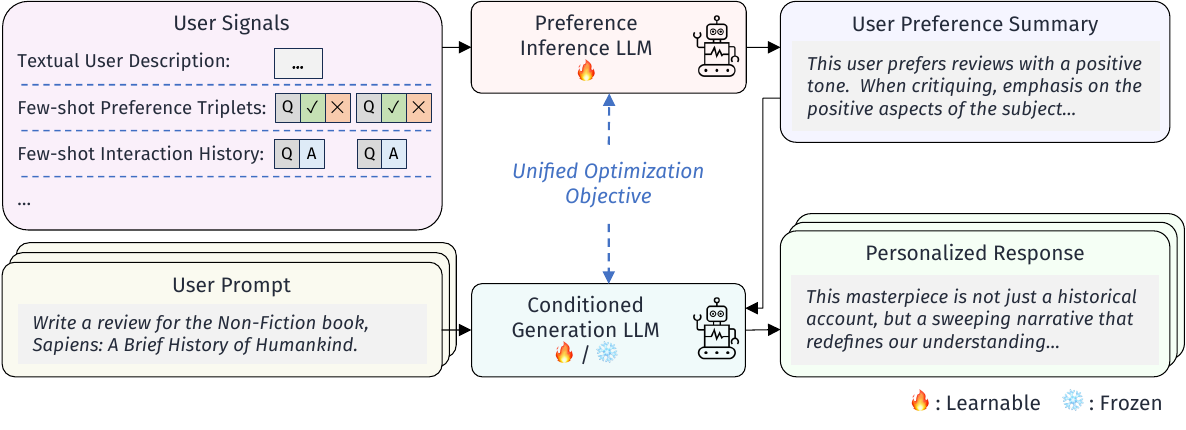}
    \caption{\textbf{Inference-time usage of \tm{}.} \tm{} instantiates the advocated modular design for user-level personalization. A shared inference model distills heterogeneous, sparse user signals into a concise natural-language preference summary, which serves as a generator-agnostic interface. A shared generator then conditions on both the task prompt and this summary to produce personalized responses. Crucially, this modular design allows inferred summaries to be reused with other frozen, off-the-shelf LLMs, including black-box commercial APIs.}
    \label{fig:usage}
\end{figure}

Large language models (LLMs) have rapidly become integral tools for information access~\citep{zhu2023large}, creativity~\citep{franceschelli2025creativity}, and decision support~\citep{vrdoljak2025review}. Despite strong aggregate performance, user satisfaction is often inconsistent: individuals may strongly prefer one model or response style over another even when benchmark results are similar~\citep{li2024personalized, guan2025survey}. Such divergence underscores a key reality: \textbf{the ``best'' model is not universal but inherently user-dependent}. The resulting challenge is not only to improve general capability but also to scalably adapt LLM behavior to diverse individual preferences.

Despite growing interest, effective LLM personalization remains technically challenging. Fine-tuning a separate model per user with alignment methods such as RLHF~\citep{ouyang2022training} or DPO~\citep{rafailov2023direct} is computationally infeasible. Recent work has therefore shifted toward shared-model approaches that condition a single generator on user-specific signals---personas, interaction histories, or preference-labeled examples, which are often scarce, heterogeneous, and noisy, making it nontrivial to extract stable preference information~\citep{singh2025fspo, zhao2023group, zhao2025nextquill, li2025million, lee2024aligning, zhou2024hypothesis}.

In this paper, we focus on \textbf{user-level personalization}, where a user's core preferences are assumed to be relatively stable across interactions. This setting has an important practical advantage: preferences need only be inferred once per user and can then be reused across many downstream prompts, making it well-suited for scalable deployment. 

User-level personalization methods vary widely in how they incorporate user signals, but they can be usefully viewed through a common lens: they decompose into a \textbf{preference inference} step that extracts stable user information from noisy signals, and a \textbf{conditioned generation} step that uses this information to produce personalized responses. In Section~\ref{sec:personalization}, we make this structure explicit and argue for a modular treatment in which these two components are separated and connected by a natural-language interface. This modular design allows the two components to be developed and improved independently by the community, and encourages a practical benefit: \textbf{generator-transferability}, where a preference summary inferred once can be reused with different generators---including black-box APIs---without retraining.

Guided by this abstraction, we propose \tm{} (\textbf{P}ersonalizing LLMs via \textbf{O}ptimized Natural Language \textbf{P}reference \textbf{I}nference), a concrete instantiation of modular user-level personalization. Figure~\ref{fig:usage} illustrates the inference-time usage of \tm{}. \tm{} employs a preference inference model to distill heterogeneous user signals into concise natural-language preference summaries, which serve as the sole interface to a generator. Both components are shared by all users and are trained under a \textbf{unified objective}, with reinforcement learning serving as an optimization tool.

We evaluate \tm{} on four benchmarks for LLM personalization, spanning sentiment analysis, explanation generation, open-ended dialogue, and forum discussion. Across these settings, \tm{} generally outperforms shared-model baselines in the non-fine-tuning regime, where optimized summaries applied to a frozen generator yield noticeable win-rate improvements over unoptimized alternatives while reducing context overhead by up to an order of magnitude. In generator-transferability experiments, the learned summaries generally transfer well to a diverse set of frozen generators---including black-box commercial APIs---lending empirical support to the modular design.

In summary, this work makes the following contributions: \textbf{(1)} We introduce \tm{}, a framework that separates user-level personalization into a preference-inference model and a conditioned generator connected by a natural-language interface, and jointly trains both components under a unified preference-optimization objective. \textbf{(2)} We show that this objective admits an information-theoretic decomposition into generator approximation error and summary informativeness, making precise how the unified objective simultaneously trains an accurate generator and an informative summarizer. \textbf{(3)} Across four benchmarks, we show that \tm{} generally improves personalization quality while reducing context overhead by up to an order of magnitude. \textbf{(4)} We demonstrate that the learned summaries transfer to frozen, off-the-shelf generators---including black-box commercial APIs. We release our implementation at \url{https://github.com/YizhuoChen99/POPI}.

\section{Related Work}

\noindent\textbf{User Profiling outside LLM Alignment.}
User modeling has a long history in NLP and related fields, spanning personalized query reformulation~\citep{zhang2022improving,hui2024rali}, summarization~\citep{zhang2024personalsum,ghodratnama2024sumrecom}, and authorship attribution~\citep{he2024authorship,deutsch2023authorship}. In recommender systems, preference inference from implicit or explicit signals has long been central, from classical matrix factorization and collaborative filtering~\citep{koren2009matrix,sarwar2001item} to recent approaches that represent user profiles directly in natural language~\citep{ramos2024transparent,zhou2024language,gao2024end,radlinski2022natural}. Related profiling techniques also appear in market research~\citep{yin2025co,rizwan2025personabot} and adaptive education~\citep{liu2025one,liu2024personality}. Our work shares both the motivation of extracting preferences from heterogeneous signals and, in particular, the choice of natural language as a user-profile representation~\citep{ramos2024transparent,radlinski2022natural,zhou2024language}. The main difference is that we optimize the natural-language summary end-to-end against a downstream preference-alignment objective, rather than treating it as a preprocessing step or a recommendation representation.

\noindent\textbf{Population-Level Preference Alignment Methods.}
Preference alignment methods aim to adapt LLMs to population-level preferences~\citep{wang2023aligning}. RLHF remains the standard approach, using reward models trained on human judgments to guide policy optimization~\citep{christiano2017deep,ouyang2022training}, with extensions that incorporate AI feedback~\citep{bai2022constitutional}. DPO~\citep{rafailov2023direct} removes explicit reward modeling, and variants such as ORPO~\citep{hong2024orpo} and SimPO~\citep{meng2024simpo} further simplify optimization. Other lines reshape the loss~\citep{ethayarajh2024kto}, enforce ranked log-probability consistency~\citep{yuan2023rrhf}, or calibrate likelihoods to preferences~\citep{zhao2023slic}. While effective for aligning to the ``average'' user, these methods largely overlook individual variation, motivating growing interest in personalized alignment~\citep{xie2025survey}.

\noindent\textbf{Personalized LLM Alignment.} Early personalization approaches adapt LLM behavior by learning user-specific adapters or embeddings, such as HyRe~\citep{lee2024testtime}, LoRe~\citep{bose2025lore}, PAL~\citep{chen2024pal}, and VPL~\citep{poddar2024personalizing}. These methods typically operate at the reward modeling level and require per-user optimization, limiting scalability. Persona-Plug~\citep{liu2025personaplug} avoids per-user optimization by learning a shared encoder that maps user histories to dense embeddings, but, like the methods above, conditions the generator through a latent interface, precluding the generator-transferability that \tm{} targets. To avoid per-user training, subsequent work adopts a meta-learning paradigm in which a shared generation model conditions on user signals provided as additional context. Representative examples include GPO~\citep{zhao2023group}, FSPO~\citep{singh2025fspo}, and NextQuill~\citep{zhao2025nextquill}, which differ in optimization objectives but share a reliance on raw or lightly processed user signals. While effective, such approaches often incur substantial context overhead and are sensitive to noisy or heterogeneous inputs. More structured methods distill user signals into explicit profiles before conditioning generation, including handcrafted persona extraction~\citep{lee2024aligning}, projection into predefined preference dimensions~\citep{li2025million}, hypothesis-based user descriptions~\citep{garbacea2025hyperalign, zhou2024hypothesis}, and zero-shot summarization or retrieval over user histories~\citep{richardson2023integrating, qiu2025measuring, salemi2024lamp}. Although these approaches improve interpretability, they typically treat preference inference as a static or heuristic preprocessing step. In contrast, \tm{} is \textbf{complementary} to this line of work: manually constructed profiles can serve as inputs to our preference inference model, which further refines them through unified optimization to produce summaries directly consumable by the generation model. Closest to \tm{}, RLPF~\citep{wu2025rlpf} and PLUS~\citep{nam2025learning} also train summarizers with reinforcement learning, but target next-item prediction in recommendation and personalized reward modeling, respectively, rather than optimizing generation directly against a preference-alignment objective.

\noindent\textbf{Prompt Optimization.} A growing body of work optimizes the text fed to LLMs to improve downstream performance~\citep{ramnath2025systematic}. Early methods search over discrete tokens via gradients~\citep{shin2020autoprompt} or reinforcement learning~\citep{deng2022rlprompt, zhang2023tempera}, while more recent approaches use LLMs themselves to propose and refine candidate instructions~\citep{zhou2023large, yang2024large, guo2024connecting, agarwal2025promptwizard}. Soft prompt tuning methods~\citep{li2021prefix, lester2021power} instead optimize continuous embeddings, which are parameter-efficient but tie the learned representation to a specific model. A few works bring prompt optimization closer to personalization: POHF~\citep{lin2024prompt} optimizes prompts from pairwise human preference feedback, and AGP~\citep{wang2025automating} optimizes user-profile-generation prompts for recommendation reranking. \tm{} can be viewed as an instance of prompt optimization in the broad sense, but differs from the works above in objective: they optimize a prompt---fixed per task, edited per query, or used to elicit user profiles---to improve performance on a target task, whereas \tm{} trains a model that maps user signals to per-user summaries under a preference-alignment objective, targeting user-level personalization and the generator-transferability it affords.

\section{Abstraction of User-level Personalization}
\label{sec:personalization}

\textbf{User-level personalization}, as the term is commonly used in the literature, refers to the task of adapting a model's generation behavior to a user's preferences, given sparse and heterogeneous user signals, under the assumption that those preferences are relatively stable across interactions and independent of the specific upcoming prompt. Preferences do evolve over time, but in many practical applications, such as personalized content generation, education, and dialogue, they remain consistent over the relevant time horizon. This stability also enables an important efficiency benefit: preferences can be inferred once and reused across many prompts, amortizing the cost of preference extraction. Many algorithms have been developed in this setting; the rest of this section makes the structure they share explicit, decomposing user-level personalization into two components, clarifying how they interact, and advocating for a modular design principle that supports generator-transferability.

\subsection{Problem Formulation and Modular Decomposition}
\label{sec:decomposition}

We consider a population of $N$ users, indexed by $i$. For each user, the available supervision takes the form of a dataset of preference-labeled examples,
\[
\mathcal{D}_i = \{(x, y_c, y_r)\},
\]
where $x$ is a prompt and $y_c, y_r$ denote the preferred and dispreferred responses, respectively. In addition, each user is associated with a set of user-specific signals carrying information about their preferences, such as personas, interaction histories, or a small number of labeled examples, which we collectively denote by $s_i$.

At an abstract level, user-level personalization methods can generally be decomposed into two components: \textbf{(1)} a \textbf{preference inference} function $f$ that maps the user signals $s_i$ to a representation of the user's preferences,
\[
z_i = f(s_i),
\]
and \textbf{(2)} a \textbf{conditioned generation} function $g$ that uses this representation, together with the prompt $x$, to produce a personalized response,
\[
y = g(x, z_i).
\]
Existing methods can be viewed as instantiations of this structure, either explicitly or implicitly, differing primarily in how $f$ and $g$ are parameterized and trained using the per-user supervision $\mathcal{D}_i$.

Making this decomposition explicit reveals a natural opportunity for \textbf{modular design}: treating preference inference and conditioned generation as separate components connected by a well-defined interface. We adopt this modular design principle for two reasons. First, it lets the two components be developed and improved independently. Second, it allows an inferred summary $z_i$ to be paired with any generator---including models with distinct architectures or those accessed only via black-box APIs---without retraining the inference model. We refer to this second property as \textbf{generator-transferability} (i.e., the learned summaries transfer across generators). It is practically valuable in deployment settings where the generator may be swapped, upgraded, or selected per query. Prior personalization work has largely under-explored this property, and we evaluate it explicitly in our experiments.

\subsection{Natural Language as a Generator-Agnostic Interface}

Given the decomposition above, the intermediate representation $z_i$ is the sole interface between preference inference and conditioned generation. If these two components are to remain meaningfully modular rather than tightly coupled, the format of this interface must be independent of any particular generator, while remaining expressive enough to capture diverse user preferences.

Natural language (with optional lightweight formatting, e.g., Markdown) is well-suited to this role. As the native input modality of modern LLMs, it is directly consumable by any model without architectural modification or fine-tuning, yet retains the expressivity needed for nuanced preferences. Latent vector representations, by contrast, tie preference inference to a specific generator; transferring them to other generators typically requires additional fine-tuning or semantic alignment.

\section{Method}\label{sec:method}

\begin{figure}[!ht]
    \centering
    \includegraphics[width=\textwidth]{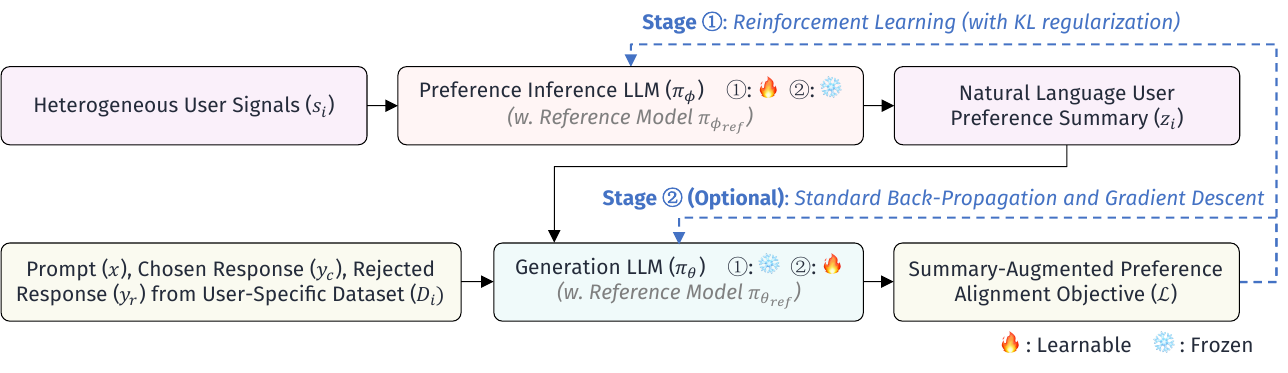}
    \caption{Training framework of \tm{}. The preference inference model $\pi_{\phi}$ maps raw user signals to a natural-language preference summary, which serves as the interface between inference and generation. Together with preference-labeled data, this summary is used to define a summary-augmented preference optimization objective that provides a unified training objective for both models. Training proceeds in two stages: $\pi_{\phi}$ is optimized with reinforcement learning, and then $\pi_{\theta}$ is optionally fine-tuned via standard backpropagation.}
    \label{fig:framework}
\end{figure}

We now instantiate the abstraction of Section~\ref{sec:personalization} and introduce \tm{}, a personalization framework that explicitly supports the modular design principle, uses natural language as the interface, and jointly trains preference inference and conditioned generation under a unified objective.

Following the decomposition, we adopt a simple parameterization for both components: preference inference $f$ is implemented as a language model $\pi_{\phi}$ that maps user-specific signals $s_i$ to a natural-language preference summary $z_i$,
\[
z_i \sim \pi_{\phi}(\cdot \mid s_i),
\]
while conditioned generation $g$ is implemented by another language model $\pi_{\theta}$ that conditions on both the task prompt $x$ and the summary $z_i$ to produce a response,
\[
y \sim \pi_{\theta}(\cdot \mid x, z_i).
\]
Both models are shared across all users. An overview of the resulting training framework is shown in Figure~\ref{fig:framework}.

\subsection{Unified Optimization Objective for Preference Inference and Generation}
\label{sec:dpo}

When user-specific preference-labeled data $\mathcal{D}_i$ is available, it is natural to use this supervision to improve both the preference inference model $\pi_{\phi}$ and the generation model $\pi_{\theta}$, rather than treating either as a fixed oracle.

\noindent\textbf{Summary-Augmented Preference Alignment Objective.}
We first derive a training objective for the generation model $\pi_{\theta}$ by adapting existing population-level preference alignment algorithms. The same adaptation procedure applies broadly across different algorithms; as a concrete illustration, we use Direct Preference Optimization (DPO)~\citep{rafailov2023direct} below. Generalization to other alignment methods is discussed in Section~\ref{sec:ipo}.

If we could afford to train a dedicated generator $\widetilde{\pi}_{\theta_i}$ for each user using $\mathcal{D}_i$, then under DPO, the per-user per-example objective would be
\[
\widetilde{\ell}_i = -\log \sigma \Bigg(
\beta \log \frac{\widetilde{\pi}_{\theta_i}(y_c \mid x)}{\pi_{\theta_\text{ref}}(y_c \mid x)}
- \beta \log \frac{\widetilde{\pi}_{\theta_i}(y_r \mid x)}{\pi_{\theta_\text{ref}}(y_r \mid x)}
\Bigg),
\]
where $\sigma$ denotes the logistic sigmoid function and $\pi_{\theta_\text{ref}}$ is a fixed reference model. Averaging over all examples and users, the aggregated loss would be
\[
\widetilde{\mathcal{L}}= \frac{1}{N} \sum_{i=1}^N
\mathbb{E}_{(x,y_c,y_r)\sim\mathcal{D}_i}
\!\left[\widetilde{\ell}_i\right].
\]
In practice, however, training a separate generator per user is infeasible. Following the decomposition introduced earlier, we therefore replace $\widetilde{\pi}_{\theta_i}(y\mid x)$ with the shared generator $\pi_{\theta}(y\mid x,z_i)$ that conditions on the user-specific natural-language summary $z_i$. This yields the \textbf{summary-augmented} DPO objective:
\begin{equation}\label{eq:sadpo}
\begin{aligned}
\ell_i
&= -\log \sigma \Bigg(
\beta \log \frac{\pi_{\theta}(y_c \mid x, z_i)}{\pi_{\theta_\text{ref}}(y_c \mid x)}
- \beta \log \frac{\pi_{\theta}(y_r \mid x, z_i)}{\pi_{\theta_\text{ref}}(y_r \mid x)}
\Bigg), \\
\mathcal{L}
&= \frac{1}{N} \sum_{i=1}^N
\mathbb{E}_{(x,y_c,y_r)\sim\mathcal{D}_i,\ z_i\sim\pi_\phi(\cdot\mid s_i)}
\!\left[\ell_i\right],
\end{aligned}
\end{equation}
which can be used to supervise the generation model $\pi_\theta$ via standard backpropagation. Note that $\mathcal{L}$ treats $z_i$ as given; we now turn to how the inference model that produces $z_i$ is itself trained.

\noindent\textbf{Supervising Preference Inference via the Same Objective.}
The summary-augmented objective $\mathcal{L}$ does more than train the generator: it can \emph{naturally serve as the training objective for the preference inference model $\pi_\phi$} as well. Intuitively, a preference summary $z_i$ receives a lower loss when conditioning on it helps $\pi_\theta$ better distinguish the preferred response from the dispreferred one.

The objective $\mathcal{L}$ is, however, \textbf{not directly differentiable} with respect to the parameters of $\pi_\phi$, because the preference summary is obtained by \textbf{sampling} $z_i \sim \pi_\phi(\cdot \mid s_i)$. Rather than modifying the objective, we treat this as a standard problem of optimizing a language model against a non-differentiable downstream loss. Any policy-gradient-based method could handle this in principle; we adopt \textbf{reinforcement learning}, specifically Group Relative Policy Optimization (GRPO)~\citep{shao2024deepseekmath}, for its efficiency and stability in fine-tuning language models~\citep{zhang2025survey,mroueh2025revisiting}.

Concretely, we view $\pi_\phi$ as a stochastic policy whose actions are natural-language preference summaries. For each user signal $s_i$, we sample a group of summaries $z_i \sim \pi_\phi(\cdot \mid s_i)$ and assign each a scalar reward equal to the negative summary-augmented objective $-\ell_i$. GRPO then optimizes $\pi_\phi$ by encouraging summaries that achieve higher relative reward within the group, so that maximizing expected reward is exactly equivalent to minimizing $\mathcal{L}$.

To further stabilize training and preserve the interpretability of summaries, we include a KL divergence regularization term that anchors $\pi_\phi$ to a fixed reference model $\pi_{\phi_\text{ref}}$, yielding the final objective:
\[
\mathcal{L}
+ \alpha\,\mathrm{KL}\!\left(
\pi_\phi(\cdot\mid s_i)\,\|\,\pi_{\phi_\text{ref}}(\cdot\mid s_i)
\right).
\]
Empirically, we observe that training with this regularized objective tends to yield \textbf{shorter summaries}, even though no length penalty appears in the objective (see Section~\ref{sec:results-elix} for a quantitative analysis). One plausible explanation is that, because only preference-predictive content contributes positively to the reward while the KL term discourages free-form deviation from the reference policy, the inference model has little incentive to retain material that is not useful for the downstream preference task, therefore producing a compression effect. This emergent behavior is practically beneficial in the personalization setting, where user signals are typically verbose and sparse, and it spares us from introducing explicit length-based constraints.

\subsection{Two-Stage Training}\label{sec:two-stage}

To preserve modularity, we adopt a two-stage training procedure rather than optimizing both models simultaneously. In Stage~1, we optimize the inference model $\pi_\phi$ using GRPO as described above, while keeping the generator $\pi_\theta$ fixed. In an \textbf{optional} Stage~2, we fine-tune $\pi_\theta$ via standard gradient-based optimization (as opposed to the RL-based optimization of $\pi_\phi$ in Stage~1) using the same objective, conditioning on the learned summaries. Algorithms~\ref{alg:popi-training} and~\ref{alg:popi-inference} in Appendix~\ref{appendix:alg} summarize the training and inference procedures.

This design involves a trade-off. By fixing the generator in Stage~1, the inference model is encouraged to produce summaries that are informative on their own and broadly useful across generators, yielding the transferability demonstrated in our experiments. However, fixing the generator may also cause the inference model to exclude information that the base generator finds difficult to exploit but that a subsequently fine-tuned generator could have leveraged, potentially limiting Stage~2 gains. We view this as a choice favoring generality and transferability, and discuss it further in the later sections.

\subsection{Generalization Across Alignment Frameworks}
\label{sec:ipo}

Although we instantiate \tm{} using DPO in Section~\ref{sec:dpo}, the framework itself can be generalized to other preference alignment algorithms, by conditioning the generator on the interface $z_i$ without altering their core structure. As another concrete illustration, we consider Identity Preference Optimization (IPO)~\citep{azar2024general}. Following the same adaptation procedure as in Section~\ref{sec:dpo}, we replace the dedicated per-user generator with the shared generator $\pi_{\theta}(y\mid x,z_i)$, yielding the \textbf{summary-augmented} IPO objective:
\begin{equation}
\begin{aligned}
\ell'_{\text{IPO},i}
&= \Bigg(
\log \frac{\pi_{\theta}(y_c \mid x, z_i)}{\pi_{\theta_\text{ref}}(y_c \mid x)}
- \log \frac{\pi_{\theta}(y_r \mid x, z_i)}{\pi_{\theta_\text{ref}}(y_r \mid x)}
- \frac{1}{2\beta}
\Bigg)^2, \\
\mathcal{L}_\text{IPO}'
&= \frac{1}{N} \sum_{i=1}^N
\mathbb{E}_{(x,y_c,y_r)\sim\mathcal{D}_i,\ z_i\sim\pi_\phi(\cdot\mid s_i)}
\!\left[\ell'_{\text{IPO},i}\right].
\end{aligned}
\end{equation}
Empirically, our ablation studies show that summary-augmented IPO achieves performance comparable to its DPO counterpart, suggesting that the framework is not tightly coupled to a specific alignment algorithm.

\subsection{A Prompt-Dependent Variant}\label{sec:prompt-dependent-variant}

\tm{} as described above targets the user-level setting, where preferences are assumed to be relatively stable across prompts. The framework nonetheless admits a natural extension beyond this scope: the preference summary $z_i$ can be additionally conditioned on the upcoming prompt $x$,
\[
z_i \sim \pi_\phi(\cdot \mid s_i, x),
\]
yielding a per-prompt summary rather than a per-user one. Because the summary must be recomputed for each new prompt, this variant falls outside the user-level scope and forfeits the modularity and reusability benefits discussed in Section~\ref{sec:personalization}. We study it as an ablation in Section~\ref{sec:ablation}. Empirically, its gains over the prompt-independent formulation are limited, so we adopt the prompt-independent variant throughout the remaining experiments.

\subsection{Interpretation: Information-Theoretic View}

To understand what $\mathcal{L}$ implicitly asks of the two models, we examine its information-theoretic structure. Building on the Bradley--Terry (BT) model of preferences~\citep{bradley1952rank} and following the analysis of~\citet{rafailov2023direct}, the summary-augmented DPO objective (Equation~\ref{eq:sadpo}) admits the decomposition
\begin{equation}\label{eq:bound}
\mathcal{L} \;=\; \underbrace{\mathrm{KL}\!\left(P(y_c \succ y_r \mid x, z_i) \,\|\, P_{\pi_\theta}(y_c \succ y_r \mid x, z_i)\right)}_{\text{generator approximation error}} \;+\; C \;-\; \underbrace{I(y_c \succ y_r \,;\, z_i \mid x)}_{\text{summary informativeness}},
\end{equation}
where $C = H(P(y_c \succ y_r \mid x))$ is independent of both models' parameters. Minimizing $\mathcal{L}$ therefore simultaneously pursues two effects: it drives the generator $\pi_\theta$ to approximate the true preference distribution under the summary, and it pushes the inference model $\pi_\phi$ to produce summaries carrying preference-relevant information beyond what the prompt $x$ already conveys. Because the mutual information is conditioned on $x$, a summary earns credit only for preference signal that the prompt itself does not supply---precisely the role the natural-language interface is meant to play in our modular design. In this sense, $\mathcal{L}$ is aligned with the two desiderata of that design: an accurate generator and an informative summary. The decomposition does not, however, split $\mathcal{L}$ into two independent subproblems---both expectations are taken over $z_i \sim \pi_\phi(\cdot \mid s_i)$, so $\pi_\phi$ shapes the generator approximation error as well, and the objective does not on its own preclude summaries tailored to the particular generator used during training. A full derivation is provided in Appendix~\ref{appendix:bound}.

\section{Experiments}

We evaluate \tm{} across four benchmarks spanning different personalization settings.

\subsection{Experimental Setup}\label{sec:setup}

\noindent\textbf{Datasets.}
We evaluate \tm{} on four public benchmarks for personalized preference alignment. The first three datasets~\citep{singh2025fspo} target distinct personalization settings: \texttt{Review} (sentiment and conciseness in media reviews), \texttt{ELIX} (explanations across five educational levels in scientific domains), and \texttt{Roleplay} (open-ended dialogue conditioned on diverse complex personas). Following prior work, we train on 20k samples per dataset, using only a small number of labeled preference pairs as user signals (four for \texttt{Review} and \texttt{ELIX}, eight for \texttt{Roleplay}) to simulate low-resource personalization. We additionally evaluate on \texttt{AlignX}~\citep{li2025million}, a large-scale forum discussion dataset. Its user signals include demographics, user-generated content, and preference-labeled pairs. \texttt{AlignX} also provides a \textbf{handcrafted 90-dimensional preference direction} derived from these signals. We therefore consider two settings: \textbf{(1)} using raw user signals as input, or \textbf{(2)} using the extracted preference directions as inputs. Following the official protocol, we train on 92k samples (7\%) and evaluate on four test splits: \texttt{DEMO}, \texttt{PAIR}, \texttt{UGC}, and \texttt{Arbitrary}, isolating different signal sources. Across all four benchmarks, each training and evaluation sample corresponds to a distinct user, so the number of users equals the dataset size in each case.

\noindent\textbf{Baselines.}
We organize baselines and our method into two groups based on whether the generation model is fine-tuned. The first group contains methods that do not fine-tune the generator.
\textbf{(1)} \textbf{Base-Model} conditions generation only on the task prompt.
\textbf{(2)} \textbf{Raw-Prompting} prepends raw user signals directly to the prompt, conceptually encompassing prior methods~\citep{zhou2024hypothesis}.
\textbf{(3)} \textbf{Summary-Prompting} conditions the frozen generator on a natural-language preference summary produced by the unoptimized inference model $\pi_{\phi_{\text{ref}}}$. This baseline serves as a controlled ablation, not as a stand-in for prior handcrafted user profile extraction methods, which are complementary to \tm{} because their handcrafted user profiles can serve as inputs to our preference inference model (an example is shown in Section~\ref{sec:alignx-results}).
\textbf{(4)} \textbf{\tm{}-Inference-Only} represents \tm{} trained with Stage~1 only. It conditions the frozen generator on \tm{}'s optimized preference summaries.
These methods can also be directly applied to personalize off-the-shelf, black-box LLMs.

The second group contains methods that additionally fine-tune the generator using preference-labeled data.
\textbf{(1)} \textbf{Raw-Aligned} fine-tunes the generator while conditioning on raw user signals, conceptually encompassing prior methods~\citep{singh2025fspo,li2025million,zhao2023group,zhao2025nextquill,lee2024aligning}.
\textbf{(2)} \textbf{Summary-Aligned} is the generator-fine-tuning counterpart of Summary-Prompting.
\textbf{(3--5)} \textbf{LoRe} and \textbf{PAL-A / PAL-B} learn \textbf{user-specific latent embeddings} that are injected into the generator during fine-tuning. As a result, these methods do not support frozen generators and do not transfer to off-the-shelf generators.
\textbf{(6)} \textbf{\tm{}-Full} corresponds to our full method described in Section~\ref{sec:method}.
For all methods, we use the same underlying preference optimization algorithm (DPO, unless otherwise specified) to ensure fair comparison.

\noindent\textbf{Evaluation Metrics.}
We evaluate personalization on \textbf{held-out users} using three complementary metrics. \textbf{Win Rate with LLM-as-a-Judge (Win Rate).} On \texttt{Review}, \texttt{ELIX}, and \texttt{Roleplay}, each test instance includes a ground-truth persona available only at evaluation time. Following the official evaluation protocol of these datasets~\citep{singh2025fspo}, we use GPT-4o~\citep{openai2024gpt4o} as a judge to compare model outputs and select the response better aligned with the persona; the presentation order of the two responses is randomized per instance to control for position bias (see Appendix~\ref{appendix:judge-prompt} for the prompting template). We further complement Win Rate with a second metric that does not rely on an external judge. \textbf{Reward Accuracy (Acc.)} measures alignment with ground-truth preference labels: a prediction is correct if the model assigns a larger log-probability margin between the preferred and dispreferred responses than the Base-Model does, capturing how much the personalization method shifts the generator toward the correct preference relative to an unpersonalized baseline. \textbf{Average Context Overhead (Avg. Len.)} reports the average number of additional tokens introduced by user-specific conditioning, measuring context efficiency.

\noindent\textbf{Hyperparameters.}
In DPO-based experiments, we scale the KL regularization weight $\alpha$ linearly with the DPO temperature $\beta$ to keep the KL and log-ratio terms on a comparable scale. Since larger $\beta$ amplifies the preference log-ratios, a proportionally larger KL weight is required to maintain balance. With this coupling, training becomes empirically robust to the specific choices of $\alpha$ (Appendix~\ref{appendix:alpha}), and we therefore fix $\alpha = 0.002 \cdot \beta$ in all experiments. For IPO-based experiments, we follow the same principle and fix $\alpha = 0.002 \cdot \tfrac{2}{\beta}$. We select $\beta$ via grid search over $\{0.1, 0.05, 0.01\}$. Unless otherwise stated, both $\pi_\phi$ and $\pi_\theta$ are initialized from LLaMA-3.2-3B-Instruct~\citep{dubey2024llama} with the default chat template. We use a batch size of 8, a learning rate of $1\times10^{-6}$, and a cosine learning rate schedule with 150 warmup steps for both training stages of \tm{} and all baselines. In Stage~1, the GRPO group size is 8 (i.e., 8 candidate summaries per user signal), the rollout count is 8, and the sampling temperature is 0.6. With batch size 8, this yields 2{,}500 RL steps for \texttt{ELIX}, \texttt{Review}, and \texttt{Roleplay} ($20{,}000/8$) and 11{,}500 steps for \texttt{AlignX} ($92{,}000/8$). All experiments use a single training seed (42).

\noindent\textbf{Generator-Transferability to Off-the-Shelf LLMs.}
To evaluate generator-transferability, we take the preference summaries learned by \tm{} and use them to condition a diverse set of frozen, off-the-shelf LLMs that were not involved in training. The set spans both open-weight and black-box commercial models: Mistral-Small (Mistral-S) and Mistral-Large (Mistral-L)~\citep{mistral2024models}, DeepSeek-R1~\citep{guo2025deepseek}, Claude-4-Sonnet (Claude-4)~\citep{anthropic2025claude4}, GPT-4o-mini~\citep{openai2024gpt4omini}, and three LLaMA-3.2-Instruct variants (Llama-1b, Llama-11b, Llama-90b)~\citep{dubey2024llama}. We use these shorthand names consistently in all result tables.

\noindent\textbf{Prompting Templates.}
Conditioned generation requires rendering structured conditioning variables as natural-language instructions that LLMs can reliably interpret. For reproducibility and fair comparison we provide the full templates used in all experiments in Appendix~\ref{appendix:prompts}, and additionally verify robustness to an alternative template set in Section~\ref{sec:ablation}.

\noindent\textbf{Evaluation Set Sizes and Confidence Intervals.}
The held-out evaluation sets contain 3{,}780 (\texttt{ELIX}), 2{,}880 (\texttt{Review}), 6{,}368 (\texttt{Roleplay}), and 3{,}716 (\texttt{AlignX}) test instances for Reward Accuracy, and 400, 320, and 400 instances for Win Rate on \texttt{ELIX}, \texttt{Review}, and \texttt{Roleplay}, respectively. All Accuracy and Win Rate values are reported with Wilson 95\% confidence intervals~\citep{wilson1927probable}. Because all experiments use a single training seed, these intervals quantify evaluation sampling error and do not capture run-to-run variance from training stochasticity. Since both metrics are defined relative to the Base-Model, its entries are fixed at 50.00 by construction. The intervals shown in those rows are therefore not uncertainty estimates for the Base-Model itself, but the width a Wilson interval attains at a proportion of 0.5 for the corresponding sample size; we report them as a reference scale for judging the intervals of the other rows.

\subsection{Results on \texttt{ELIX}, \texttt{Review}, and \texttt{Roleplay}}
\label{sec:results-elix}

We evaluate \tm{} on \texttt{ELIX}, \texttt{Review}, and \texttt{Roleplay}, where ground-truth personas are used \emph{only} at evaluation time for LLM-as-a-judge comparisons and are never observed during personalized generation.

\noindent\textbf{Personalization Performance and Efficiency.}
Table~\ref{tab:elix-main} reports main results in two regimes. Without generator fine-tuning, \tm{}-Inference-Only outperforms Raw-Prompting and Summary-Prompting on all three datasets, with the largest gains on \texttt{Review} (win rate 80.99 vs.\ 57.65) and \texttt{ELIX} (63.84 vs.\ 49.31), and a smaller but consistent improvement on \texttt{Roleplay} (55.36 vs.\ 39.39)---all at roughly an order-of-magnitude reduction in context overhead (e.g., $\sim$53 vs.\ $\sim$3{,}175 tokens on \texttt{ELIX}). On all three datasets, its Win Rate interval is disjoint from those of both prompting baselines. With generator fine-tuning, \tm{}-Full achieves the highest Win Rate on \texttt{ELIX} and \texttt{Roleplay}, though its interval narrowly overlaps those of Raw-Aligned on \texttt{ELIX} and LoRe on \texttt{Roleplay}; on \texttt{Review}, Raw-Aligned achieves a marginally higher win rate (89.54 vs.\ 88.08, with overlapping intervals) at $\sim$10$\times$ the context cost. LoRe and PAL-A/B minimize context overhead via learned per-user embeddings, but couple representations to a specific generator, precluding the transferability evaluated below.

\begin{table}[!ht]
\centering
\caption{Performance comparison on \texttt{ELIX}, \texttt{Review}, and \texttt{Roleplay} using Avg. Len., Acc., and Win Rate. All Acc.\ and Win Rate values are reported with Wilson 95\% confidence intervals~\citep{wilson1927probable}. $\dagger$ indicates a \textbf{special token} that is mapped to a learned user-specific latent embedding, rather than a regular token from the model's original tokenizer vocabulary.}
\begin{adjustbox}{max width=\linewidth}
\small
\begin{tabular}{lrrrrrrrrr}
\toprule
\multirow{2}{*}{Method} & \multicolumn{3}{c}{\texttt{ELIX}} & \multicolumn{3}{c}{\texttt{Review}} & \multicolumn{3}{c}{\texttt{Roleplay}}\\
\cmidrule(lr){2-4} \cmidrule(lr){5-7} \cmidrule(lr){8-10}
& Avg. Len. $\downarrow$ & Acc. (\%) & Win Rate (\%) & Avg. Len. $\downarrow$ & Acc. (\%) & Win Rate (\%) & Avg. Len. $\downarrow$ & Acc. (\%) & Win Rate (\%) \\
\midrule
Base-Model & \multicolumn{1}{r}{--} & 50.00\ci{1.59} & 50.00\ci{4.88} & \multicolumn{1}{r}{--} & 50.00\ci{1.82} & 50.00\ci{5.45} & \multicolumn{1}{r}{--} & 50.00\ci{1.23} & 50.00\ci{4.88} \\
\midrule
Raw-Prompting & 3175.08 & 56.12\ci{1.58} & 49.31\ci{4.88} & 3049.10 & 51.08\ci{1.82} & 53.46\ci{5.43} & 3020.61 & 53.06\ci{1.23} & 35.26\ci{4.66} \\
Summary-Prompting & 536.41 & 55.61\ci{1.58} & 47.40\ci{4.87} & 535.44 & 51.91\ci{1.82} & 57.65\ci{5.38} & 529.47 & 53.28\ci{1.23} & 39.39\ci{4.77} \\
\rowcolor{myrose}
\tm{}-Inference-Only & \textbf{52.52} & \textbf{71.48}\ci{1.44} & \textbf{63.84}\ci{4.69} & \textbf{320.80} & \textbf{91.81}\ci{1.00} & \textbf{80.99}\ci{4.29} & \textbf{200.22} & \textbf{54.81}\ci{1.22} & \textbf{55.36}\ci{4.85} \\
\midrule
Raw-Aligned & 3175.08 & 73.94\ci{1.40} & 56.14\ci{4.84} & 3049.10 & 95.07\ci{0.79} & \textbf{89.54}\ci{3.37} & 3020.61 & 65.31\ci{1.17} & 38.66\ci{4.75} \\
Summary-Aligned & 536.41 & 61.57\ci{1.55} & 40.41\ci{4.79} & 535.44 & 77.64\ci{1.52} & 65.24\ci{5.19} & 529.47 & 66.14\ci{1.16} & 53.45\ci{4.87} \\
LoRe & \textit{1.00} $\dagger$ & 63.27\ci{1.54} & 51.03\ci{4.88} & \textit{1.00} $\dagger$ & 83.54\ci{1.35} & 67.76\ci{5.10} & \textit{1.00} $\dagger$ & 66.77\ci{1.16} & 65.29\ci{4.65} \\
PAL-A & \textit{1.00} $\dagger$ & 61.20\ci{1.55} & 51.28\ci{4.88} & \textit{1.00} $\dagger$ & 91.22\ci{1.03} & 78.69\ci{4.47} & \textit{1.00} $\dagger$ & 66.60\ci{1.16} & 28.41\ci{4.40} \\
PAL-B & \textit{1.00} $\dagger$ & 64.80\ci{1.52} & 51.68\ci{4.87} & \textit{1.00} $\dagger$ & 92.54\ci{0.96} & 78.44\ci{4.49} & \textit{1.00} $\dagger$ & 66.94\ci{1.16} & 33.04\ci{4.59} \\
\rowcolor{myrose}
\tm{}-Full & \textbf{52.52} & \textbf{80.14}\ci{1.27} & \textbf{63.97}\ci{4.68} & \textbf{320.80} & \textbf{95.76}\ci{0.74} & 88.08\ci{3.56} & \textbf{200.22} & \textbf{72.36}\ci{1.10} & \textbf{70.12}\ci{4.47} \\
\bottomrule
\end{tabular}
\end{adjustbox}
\label{tab:elix-main}
\end{table}

\noindent\textbf{Relationship Between Accuracy and Win Rate.}
The two metrics probe personalization differently: Accuracy measures how far a method shifts log-probability margins toward the preferred response, while Win Rate measures how an external judge rates the text actually generated. Across all 27 method-dataset combinations in Table~\ref{tab:elix-main}, the Pearson correlation between them is 0.80, indicating that the two metrics largely agree without being redundant. Generator fine-tuning improves Accuracy more consistently than Win Rate, raising the former in all nine method pairs and the latter in eight. This follows from what each metric measures: fine-tuning directly optimizes the log-probability margins Accuracy is computed from, whereas its effect on judged generation quality is indirect.

\begin{figure}[!ht]
    \centering
    \includegraphics[width=0.6\linewidth]{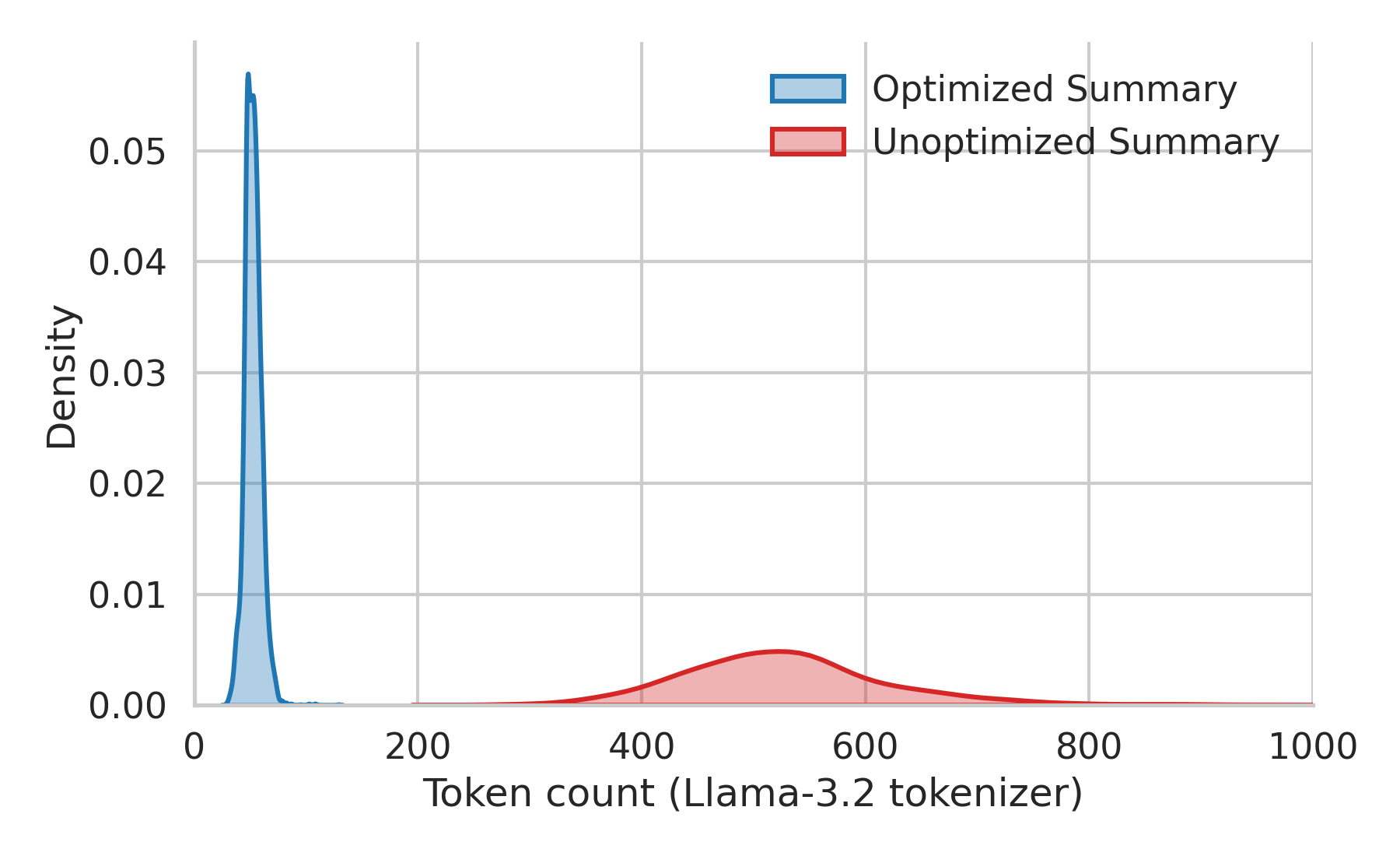}
    \caption{Distribution of preference summary lengths (in tokens) on \texttt{ELIX}, before and after optimization. The optimized inference model produces substantially shorter summaries, concentrating around $\sim$50 tokens compared to $\sim$500 for the unoptimized reference model.}
    \label{fig:summary-length}
\end{figure}

\noindent\textbf{Analysis of Learned Summaries.}
Figure~\ref{fig:summary-length} compares summary lengths before and after optimization on \texttt{ELIX}: the unoptimized reference model produces summaries spanning roughly 400--700 tokens, while the optimized model concentrates sharply around 50 tokens, confirming the emergent compression discussed in Section~\ref{sec:dpo}. We also verify that the optimized summaries remain linguistically well-formed rather than degenerating into adversarial triggers: \textbf{99.8\% of tokens} are covered by the top 50{,}000 most frequent English words in the SUBTLEX-US frequency corpus (compared with 98.8\% for the Brown Corpus), with no tokens with $\geq 3$ repeated characters (e.g., \texttt{aaa}), no symbol-heavy tokens (e.g., \texttt{a\&\&}), no consonant-only tokens of length $\geq 3$ (e.g., \texttt{qwty}), and no tokens exceeding 30 characters.

\noindent\textbf{Length-Matched Baseline.}
To test whether \tm{}'s gains stem from the \emph{content} of the optimized summaries rather than from length reduction alone, we construct a length-matched Summary-Prompting baseline on \texttt{ELIX} by instructing the unoptimized inference model to produce short summaries matching \tm{}'s output format. Table~\ref{tab:length-matched} shows that compressing unoptimized summaries to comparable length (78.74 tokens) leaves performance essentially unchanged (47.54 vs.\ 47.40 Win Rate; 55.83 vs.\ 55.61 Acc.), while \tm{}-Inference-Only remains substantially ahead on both metrics at an even shorter length. This confirms that the improvements derive from optimized summary content rather than brevity or formatting.

\begin{table}[!ht]
\caption{Comparison against a length-matched Summary-Prompting baseline on \texttt{ELIX} using Avg. Len., Acc., and Win Rate.}
\centering
\begin{adjustbox}{max width=0.75\linewidth}
\small
\begin{tabular}{lrrr}
\toprule
\multirow{2}{*}{Method} & \multicolumn{3}{c}{\texttt{ELIX}} \\\cmidrule{2-4}
& Avg. Len. $\downarrow$ & Acc. (\%) & Win Rate (\%) \\
\midrule
Base-Model & \multicolumn{1}{r}{--} & 50.00\ci{1.59} & 50.00\ci{4.88} \\
\midrule
Summary-Prompting & 536.41 & 55.61\ci{1.58} & 47.40\ci{4.87} \\
Summary-Prompting (Length-Matched) & 78.74 & 55.83\ci{1.58} & 47.54\ci{4.87} \\
\rowcolor{myrose}
\tm{}-Inference-Only & \textbf{52.52} & \textbf{71.48}\ci{1.44} & \textbf{63.84}\ci{4.69} \\
\bottomrule
\end{tabular}
\end{adjustbox}
\label{tab:length-matched}
\end{table}

\noindent\textbf{Generator-Transferability.}
Tables~\ref{tab:elix-transfer}--\ref{tab:roleplay-transfer} evaluate whether optimized summaries remain useful when paired with frozen, off-the-shelf LLMs. On \texttt{Review}, \tm{}-Inference-Only outperforms all baselines on every tested generator, averaging 79.88\% win rate versus 62.92\% (Raw-Prompting) and 54.26\% (Summary-Prompting). On \texttt{ELIX}, the pattern is similar (70.77\% vs.\ 63.11\% and 55.79\%). Notably, \tm{}-Inference-Only, whose 3B inference model is optimized by \tm{}, outperforms Summary-Prompting (Llama-90b), whose summaries come from an unoptimized 90B model (70.77\% vs.\ 63.98\% on average). Raw-Prompting sometimes underperforms despite access to more information, suggesting that unfiltered signals introduce noise that outweighs their informational content. On \texttt{Roleplay}, results are more mixed: \tm{}-Inference-Only averages 52.61\% win rate---a modest improvement over the 50\% Base-Model baseline and above the below-50\% averages of Raw-Prompting (43.02\%) and Summary-Prompting (47.60\%), but with substantial variation across generators. We attribute this to the low signal-to-noise ratio in Roleplay's user signals: each user is represented by only eight preference pairs spanning diverse topics, with limited stylistic or topical overlap with the downstream generation. Prepending these user signals directly introduces more noise than useful guidance, actively perturbing downstream generation rather than improving it. \tm{}'s optimization-driven compression mitigates this by partially filtering out non-predictive content, though the performance gains remain modest when the underlying signals carry limited extractable preference information.

\newcommand{\heatmap}[1]{%
  \cellcolor{myrose!\fpeval{(#1-18.75)*100/63.26}!myblue} #1%
}

\begin{table}[!htbp]
\caption{Generator-transferability on \texttt{ELIX}: Win Rate (\%) across personalization methods applied to frozen, off-the-shelf LLMs ($N=200$ per generator). Wilson 95\% CIs are reported per generator; they are omitted for the averaged column because shared test prompts across generators violate the independence assumption.$^\ast$ Rows marked (Prompt-Dep.) use the prompt-dependent variant of Section~\ref{sec:prompt-dependent-variant}; (Llama-90b) denotes Summary-Prompting with summaries produced by an unoptimized Llama-90b instead of the reference 3B model; (IPO Based) uses the summary-augmented IPO objective (Section~\ref{sec:ipo}).}
\centering
\begin{adjustbox}{max width=\linewidth}
\small
\begin{tabular}{lcccccccccc}
\toprule
\multirow{2}{*}{Method} & \multicolumn{8}{c}{Win Rate (\%) on \texttt{ELIX}} & \multirow{2}{*}{Avg.$^\ast$} \\\cmidrule{2-9}
& Mistral-S & Mistral-L & DeepSeek-R1 & Llama-1b & Llama-11b & Llama-90b & Claude-4 & GPT-4o-mini &   \\
\midrule
Base-Model & \heatmap{50.00}\ci{6.93} & \heatmap{50.00}\ci{6.93} & \heatmap{50.00}\ci{6.93} & \heatmap{50.00}\ci{6.93} & \heatmap{50.00}\ci{6.93} & \heatmap{50.00}\ci{6.93} & \heatmap{50.00}\ci{6.93} & \heatmap{50.00}\ci{6.93} & \heatmap{50.00} \\
Raw-Prompting       & \heatmap{67.27}\ci{6.50} & \heatmap{72.16}\ci{6.21} & \heatmap{79.38}\ci{5.61} & \heatmap{27.16}\ci{6.16} & \heatmap{56.05}\ci{6.88} & \heatmap{59.03}\ci{6.82} & \heatmap{72.38}\ci{6.20} & \heatmap{71.43}\ci{6.26} & \heatmap{63.11} \\
Summary-Prompting                  & \heatmap{61.13}\ci{6.76} & \heatmap{67.15}\ci{6.51} & \heatmap{74.53}\ci{6.04} & \heatmap{18.75}\ci{5.41} & \heatmap{48.64}\ci{6.93} & \heatmap{48.33}\ci{6.93} & \heatmap{59.96}\ci{6.79} & \heatmap{67.85}\ci{6.47} & \heatmap{55.79} \\
Summary-Prompting (Prompt-Dep.) & \heatmap{55.73}\ci{6.88} & \heatmap{51.91}\ci{6.92} & \heatmap{66.65}\ci{6.53} & \heatmap{43.66}\ci{6.87} & \heatmap{41.47}\ci{6.83} & \heatmap{44.37}\ci{6.89} & \heatmap{58.00}\ci{6.84} & \heatmap{57.38}\ci{6.85} & \heatmap{52.40} \\
Summary-Prompting (Llama-90b)   & \heatmap{68.36}\ci{6.45} & \heatmap{75.09}\ci{5.99} & \heatmap{73.51}\ci{6.12} & \heatmap{47.25}\ci{6.92} & \heatmap{55.60}\ci{6.89} & \heatmap{55.15}\ci{6.89} & \heatmap{69.39}\ci{6.39} & \heatmap{67.50}\ci{6.49} & \heatmap{63.98} \\
\rowcolor{myrose}
\tm{}-Inference-Only  & \textbf{\heatmap{81.23}}\ci{5.41} & \textbf{\heatmap{81.50}}\ci{5.38} & \heatmap{70.88}\ci{6.30} & \textbf{\heatmap{62.65}}\ci{6.70} & \textbf{\heatmap{63.53}}\ci{6.67} & \heatmap{59.62}\ci{6.80} & \heatmap{70.22}\ci{6.34} & \heatmap{76.53}\ci{5.87} & \heatmap{70.77} \\
\rowcolor{myrose}
\tm{}-Inference-Only  (Prompt-Dep.) & \heatmap{79.50}\ci{5.60} & \heatmap{76.70}\ci{5.86} & \textbf{\heatmap{82.01}}\ci{5.32} & \heatmap{61.55}\ci{6.74} & \heatmap{61.89}\ci{6.73} & \textbf{\heatmap{63.08}}\ci{6.69} & \textbf{\heatmap{76.29}}\ci{5.89} & \heatmap{75.84}\ci{5.93} & \textbf{\heatmap{72.11}} \\
\rowcolor{myrose}
\tm{}-Inference-Only  (IPO Based) & \heatmap{78.05}\ci{5.74} & \heatmap{80.57}\ci{5.48} & \heatmap{71.75}\ci{6.24} & \heatmap{59.78}\ci{6.80} & \heatmap{62.79}\ci{6.70} & \heatmap{60.99}\ci{6.76} & \heatmap{68.23}\ci{6.45} & \textbf{\heatmap{80.83}}\ci{5.46} & \heatmap{70.37} \\
\bottomrule
\end{tabular}
\end{adjustbox}
\label{tab:elix-transfer}
\end{table}

\renewcommand{\heatmap}[1]{%
  \cellcolor{myrose!\fpeval{(#1 - 34.82)*100/50.85}!myblue} #1%
}
\begin{table}[!htbp]
\caption{Generator-transferability on \texttt{Review}: Win Rate (\%) across personalization methods applied to frozen, off-the-shelf LLMs ($N=160$ per generator). Wilson 95\% CIs are reported per generator; they are omitted for the averaged column.$^\ast$}
\centering
\begin{adjustbox}{max width=\linewidth}
\small
\begin{tabular}{lccccccccc}
\toprule
\multirow{2}{*}{Method} & \multicolumn{8}{c}{Win Rate (\%) on \texttt{Review}} & \multirow{2}{*}{Avg.$^\ast$} \\\cmidrule{2-9}
& Mistral-S & Mistral-L & DeepSeek-R1 & Llama-1b & Llama-11b & Llama-90b & Claude-4 & GPT-4o-mini &   \\
\midrule
Base-Model & \heatmap{50.00}\ci{7.75} & \heatmap{50.00}\ci{7.75} & \heatmap{50.00}\ci{7.75} & \heatmap{50.00}\ci{7.75} & \heatmap{50.00}\ci{7.75} & \heatmap{50.00}\ci{7.75} & \heatmap{50.00}\ci{7.75} & \heatmap{50.00}\ci{7.75} & \heatmap{50.00} \\
Raw-Prompting & \heatmap{48.66}\ci{7.74} & \heatmap{58.74}\ci{7.63} & \heatmap{79.78}\ci{6.22} & \heatmap{34.82}\ci{7.38} & \heatmap{67.87}\ci{7.24} & \heatmap{70.88}\ci{7.04} & \heatmap{70.80}\ci{7.05} & \heatmap{71.84}\ci{6.97} & \heatmap{62.92} \\
Summary-Prompting & \heatmap{47.70}\ci{7.74} & \heatmap{52.80}\ci{7.74} & \heatmap{61.88}\ci{7.53} & \heatmap{40.32}\ci{7.60} & \heatmap{56.66}\ci{7.68} & \heatmap{58.10}\ci{7.65} & \heatmap{54.70}\ci{7.71} & \heatmap{61.94}\ci{7.52} & \heatmap{54.26} \\
\rowcolor{myrose}
\tm{}-Inference-Only & \textbf{\heatmap{72.51}}\ci{6.92} & \textbf{\heatmap{85.67}}\ci{5.43} & \textbf{\heatmap{84.90}}\ci{5.55} & \textbf{\heatmap{77.28}}\ci{6.49} & \textbf{\heatmap{83.53}}\ci{5.75} & \textbf{\heatmap{79.81}}\ci{6.22} & \textbf{\heatmap{74.08}}\ci{6.79} & \textbf{\heatmap{81.27}}\ci{6.05} & \textbf{\heatmap{79.88}} \\
\bottomrule
\end{tabular}
\end{adjustbox}
\label{tab:review-transfer}
\end{table}

\renewcommand{\heatmap}[1]{%
  \cellcolor{myrose!\fpeval{(#1 - 13.99)*100/56.35}!myblue} #1%
}
\begin{table}[!htbp]
\caption{Generator-transferability on \texttt{Roleplay}: Win Rate (\%) across personalization methods applied to frozen, off-the-shelf LLMs ($N=200$ per generator). Wilson 95\% CIs are reported per generator; they are omitted for the averaged column.$^\ast$}
\centering
\begin{adjustbox}{max width=\linewidth}
\small
\begin{tabular}{lccccccccc}
\toprule
\multirow{2}{*}{Method} & \multicolumn{8}{c}{Win Rate (\%) on \texttt{Roleplay}} & \multirow{2}{*}{Avg.$^\ast$} \\\cmidrule{2-9}
& Mistral-S & Mistral-L & DeepSeek-R1 & Llama-1b & Llama-11b & Llama-90b & Claude-4 & GPT-4o-mini &   \\
\midrule
Base-Model & \heatmap{50.00}\ci{6.93} & \heatmap{50.00}\ci{6.93} & \heatmap{50.00}\ci{6.93} & \heatmap{50.00}\ci{6.93} & \textbf{\heatmap{50.00}}\ci{6.93} & \heatmap{50.00}\ci{6.93} & \textbf{\heatmap{50.00}}\ci{6.93} & \heatmap{50.00}\ci{6.93} & \heatmap{50.00} \\
Raw-Prompting & \heatmap{66.78}\ci{6.53} & \textbf{\heatmap{70.34}}\ci{6.33} & \heatmap{47.74}\ci{6.92} & \heatmap{23.74}\ci{5.90} & \heatmap{34.75}\ci{6.60} & \heatmap{34.62}\ci{6.59} & \heatmap{13.99}\ci{4.81} & \heatmap{52.19}\ci{6.92} & \heatmap{43.02} \\
Summary-Prompting & \textbf{\heatmap{68.62}}\ci{6.43} & \heatmap{66.31}\ci{6.55} & \heatmap{58.18}\ci{6.84} & \heatmap{22.41}\ci{5.78} & \heatmap{31.25}\ci{6.42} & \heatmap{39.85}\ci{6.79} & \heatmap{40.65}\ci{6.81} & \heatmap{53.49}\ci{6.91} & \heatmap{47.60} \\
\rowcolor{myrose}
\tm{}-Inference-Only & \heatmap{47.89}\ci{6.92} & \heatmap{57.52}\ci{6.85} & \textbf{\heatmap{62.93}}\ci{6.69} & \textbf{\heatmap{56.39}}\ci{6.87} & \heatmap{48.17}\ci{6.93} & \textbf{\heatmap{51.27}}\ci{6.93} & \heatmap{37.57}\ci{6.71} & \textbf{\heatmap{59.12}}\ci{6.81} & \textbf{\heatmap{52.61}} \\
\bottomrule
\end{tabular}
\end{adjustbox}
\label{tab:roleplay-transfer}
\end{table}

\noindent\textbf{Qualitative Case Study.}
Figures~\ref{fig:summary-case} and~\ref{fig:case} (Appendices~\ref{appendix:summary-case} and~\ref{appendix:case}) show representative examples on \texttt{ELIX}. Before optimization, summaries are verbose and structurally generic across personas; after optimization, they are compact ($\sim$50 tokens) and differentiated---for instance, the learned summary includes ``child-centric language and simple definitions'' for a held-out user with the child ground-truth persona, versus ``technical accuracy and technical jargon'' for another held-out user with the expert ground-truth persona. These summaries guide generation toward responses aligned with users' preferences. Appendix~\ref{appendix:case-cross} shows corresponding examples on \texttt{Review} and \texttt{Roleplay}. On \texttt{Review}, the optimized summary aligns closely with the ground-truth persona, recovering the user's preference for positive reviews where the unoptimized summary instead describes generic qualities of a well-written review. On \texttt{Roleplay}, the ground-truth personas are detailed, multi-faceted descriptions of synthetic individuals, while the eight preference pairs serving as user signals are noisy, span a wide range of topics, and often bear little direct relation to the persona dimensions that matter for evaluation. The optimized summary consequently distills these signals into a few sentences about the user's general topical interests---not strongly indicative for generation, but not actively harmful either---consistent with the modest improvement over baselines observed on \texttt{Roleplay}.

\subsection{Results on \texttt{AlignX} Dataset}
\label{sec:alignx-results}

We next evaluate \tm{} on \texttt{AlignX}, a large-scale forum discussion benchmark that provides two types of user signals (Section~\ref{sec:setup}). We report only Reward Accuracy on this benchmark: Win Rate requires a held-out ground-truth persona for the LLM judge to assess which response better matches the user, and \texttt{AlignX} supplies only preference-labeled pairs. Tables~\ref{tab:alignx} and~\ref{tab:alignx-90d} summarize results across both signal types and four official test splits. Without generator fine-tuning, \tm{}-Inference-Only outperforms Raw-Prompting and Summary-Prompting on every split and signal type, while generally using far fewer context tokens. Relative to the Base-Model, the gains are significant on \texttt{Arbitrary} and \texttt{DEMO}, where user signals carry the most preference-relevant information, but not on \texttt{PAIR} and \texttt{UGC}.

When generator fine-tuning is allowed, \tm{}-Full generally reduces context overhead by a large margin relative to Raw-Aligned, but trails it in accuracy moderately on most splits. A likely contributing factor is the two-stage training design discussed in Section~\ref{sec:two-stage}: because the inference model is optimized against a frozen base generator in Stage~1, it may compress away signal that the base generator cannot exploit but that a subsequently fine-tuned generator could have leveraged. We regard closing this gap---for example, through constrained iterative co-training of both components (Section~\ref{sec:ablation})---as an important direction for future work.

\noindent\textbf{\tm{} is complementary to handcrafted user profile extraction methods.} The 90-dimensional preference direction provided with \texttt{AlignX} is itself a carefully engineered representation of user preferences~\citep{li2025million}. When used as input to \tm{}'s inference model (Table~\ref{tab:alignx-90d}), the optimized summaries improve accuracy on every split, significantly so on \texttt{Arbitrary} and \texttt{DEMO} (e.g., 56.55\% vs.\ 51.97\% on \texttt{Arbitrary}), though the intervals overlap on \texttt{PAIR} and \texttt{UGC}, while compressing context (e.g., from $\sim$260 to $\sim$91 tokens on \texttt{Arbitrary}). This suggests that \tm{}'s optimization can serve as a refinement layer on top of existing handcrafted user profile extraction pipelines, rather than as a replacement.

\begin{table}[!htbp]
\caption{Performance Comparisons on \texttt{AlignX} using Avg. Len.\ and Acc., across four official test splits (Arbitrary, DEMO, PAIR, UGC), with Wilson 95\% CIs ($N=3{,}716$). User signals are the original heterogeneous user signals provided with \texttt{AlignX}.}
\centering
\begin{adjustbox}{max width=\linewidth}
\small
\begin{tabular}{lrrrrrrrr}
\toprule
\multirow{3}{*}{Method} & \multicolumn{8}{c}{\texttt{AlignX} (with Original Heterogeneous User Signals)} \\\cmidrule(lr){2-9}
& \multicolumn{2}{c}{Arbitrary} & \multicolumn{2}{c}{DEMO} & \multicolumn{2}{c}{PAIR} & \multicolumn{2}{c}{UGC} \\\cmidrule(lr){2-3} \cmidrule(lr){4-5} \cmidrule(lr){6-7} \cmidrule(lr){8-9}
& Avg. Len. & Acc. & Avg. Len. & Acc. & Avg. Len. & Acc. & Avg. Len. & Acc.\\
\midrule
Base-Model & \multicolumn{1}{r}{--} & 50.00\ci{1.61} & \multicolumn{1}{r}{--} & 50.00\ci{1.61} & \multicolumn{1}{r}{--} & 50.00\ci{1.61} & \multicolumn{1}{r}{--} & 50.00\ci{1.61} \\
\midrule
Raw-Prompting & 1402.86 & 51.89\ci{1.61} & \textbf{105.09} & 58.27\ci{1.58} & 1501.26 & 49.33\ci{1.61} & 1048.37 & 49.06\ci{1.61} \\
Summary-Prompting & 288.75 & 52.10\ci{1.61} & 138.35 & 57.73\ci{1.59} & 315.74 & 48.63\ci{1.61} & 284.24 & 49.60\ci{1.61} \\
\rowcolor{myrose}
\tm{}-Inference-Only  & \textbf{151.65} & \textbf{54.63}\ci{1.60} & 188.25 & \textbf{60.37}\ci{1.57} & \textbf{128.04} & \textbf{50.22}\ci{1.61} & \textbf{134.94} & \textbf{50.92}\ci{1.61} \\
\midrule
Raw-Aligned & 1402.86 & \textbf{74.60}\ci{1.40} & \textbf{105.09} & \textbf{92.51}\ci{0.85} & 1501.26 & \textbf{55.95}\ci{1.60} & 1048.37 & \textbf{57.36}\ci{1.59} \\
Summary-Aligned & 288.75 & 69.56\ci{1.48} & 138.35 & 89.79\ci{0.97} & 315.74 & 51.70\ci{1.61} & 284.24 & 51.99\ci{1.61} \\
\rowcolor{myrose}
\tm{}-Full  & \textbf{151.65} & 71.12\ci{1.46} & 188.25 & 90.44\ci{0.95} & \textbf{128.04} & 52.51\ci{1.60} & \textbf{134.94} & 54.45\ci{1.60} \\
\bottomrule
\end{tabular}
\end{adjustbox}
\label{tab:alignx}
\end{table}

\begin{table}[!htbp]
\caption{Performance Comparisons on \texttt{AlignX} using Avg. Len.\ and Acc., across four official test splits (Arbitrary, DEMO, PAIR, UGC), with Wilson 95\% CIs ($N=3{,}716$). User signals are the 90-dimensional preference directions extracted by the \texttt{AlignX} authors.}
\centering
\begin{adjustbox}{max width=\linewidth}
\small
\begin{tabular}{lrrrrrrrr}
\toprule
\multirow{3}{*}{Method} & \multicolumn{8}{c}{\texttt{AlignX} (with 90-Dimensional Preference Directions as User Signals)} \\\cmidrule(lr){2-9}
& \multicolumn{2}{c}{Arbitrary} & \multicolumn{2}{c}{DEMO} & \multicolumn{2}{c}{PAIR} & \multicolumn{2}{c}{UGC} \\\cmidrule(lr){2-3} \cmidrule(lr){4-5} \cmidrule(lr){6-7} \cmidrule(lr){8-9}
& Avg. Len. & Acc. & Avg. Len. & Acc. & Avg. Len. & Acc. & Avg. Len. & Acc.\\
\midrule
Base-Model & \multicolumn{1}{r}{--} & 50.00\ci{1.61} & \multicolumn{1}{r}{--} & 50.00\ci{1.61} & \multicolumn{1}{r}{--} & 50.00\ci{1.61} & \multicolumn{1}{r}{--} & 50.00\ci{1.61} \\
\midrule
Raw-Prompting & 259.92 & 51.97\ci{1.61} & 117.94 & 54.90\ci{1.60} & 233.88 & 49.81\ci{1.61} & 184.68 & 51.19\ci{1.61} \\
Summary-Prompting & 257.52 & 52.69\ci{1.60} & 186.25 & 55.79\ci{1.60} & 248.22 & 50.62\ci{1.61} & 239.68 & 50.94\ci{1.61} \\
\rowcolor{myrose}
\tm{}-Inference-Only  & \textbf{91.47} & \textbf{56.55}\ci{1.59} & \textbf{72.83} & \textbf{62.72}\ci{1.55} & \textbf{86.30} & \textbf{52.80}\ci{1.60} & \textbf{87.97} & \textbf{52.67}\ci{1.60} \\
\midrule
Raw-Aligned & 259.92 & \textbf{74.41}\ci{1.40} & 117.94 & \textbf{87.88}\ci{1.05} & 233.88 & \textbf{59.78}\ci{1.58} & 184.68 & \textbf{58.76}\ci{1.58} \\
Summary-Aligned & 257.52 & 70.56\ci{1.46} & 186.25 & 84.24\ci{1.17} & 248.22 & 58.86\ci{1.58} & 239.68 & 58.54\ci{1.58} \\
\rowcolor{myrose}
\tm{}-Full  & \textbf{91.47} & 69.45\ci{1.48} & \textbf{72.83} & 84.19\ci{1.17} & \textbf{86.30} & 58.73\ci{1.58} & \textbf{87.97} & 58.54\ci{1.58} \\
\bottomrule
\end{tabular}
\end{adjustbox}
\label{tab:alignx-90d}
\end{table}

\subsection{Ablation Studies}
\label{sec:ablation}

We conduct six ablation studies on the \texttt{ELIX} dataset to better understand the design choices of \tm{}: \textbf{(1)} the effect of using IPO as the underlying framework, \textbf{(2)} the effect of prompt-dependent summaries, \textbf{(3)} two-stage vs.\ iterative co-training, \textbf{(4)} sensitivity to the scaling parameter $\beta$ (Appendix~\ref{appendix:beta}), \textbf{(5)} sensitivity to the KL regularization weight $\alpha$ (Appendix~\ref{appendix:alpha}), and \textbf{(6)} robustness to prompting templates (Appendix~\ref{appendix:template}).

\noindent\textbf{IPO as the Underlying Framework.}
In this study, we adopt IPO as the underlying preference alignment framework for \tm{} and baselines, instead of DPO, as described in Section~\ref{sec:ipo}. Main results are shown in Table~\ref{tab:elix-ipo}, with transfer results reported in Table~\ref{tab:elix-transfer}. Across both settings, \tm{} based on IPO performs comparably to DPO across all metrics, confirming that \tm{} is broadly compatible with different underlying alignment frameworks.

\begin{table}[!ht]
    \caption{Performance comparisons on \texttt{ELIX} using Avg. Len., Acc., and Win Rate, with the summary-augmented IPO objective.}
   \centering
   \begin{adjustbox}{max width=0.55\linewidth}
   \small
   \begin{tabular}{lrrr}
   \toprule
   \multirow{2}{*}{Method} & \multicolumn{3}{c}{\texttt{ELIX} (IPO Based)} \\\cmidrule(lr){2-4}
   & Avg. Len. $\downarrow$ & Acc. (\%) & Win Rate (\%) \\
   \midrule
   Base-Model & \multicolumn{1}{r}{--} & 50.00\ci{1.59} & 50.00\ci{4.88} \\
   \midrule
   Raw-Prompting & 3175.08 & 56.12\ci{1.58} & 49.31\ci{4.88} \\
   Summary-Prompting & 536.41 & 55.61\ci{1.58} & 47.40\ci{4.87} \\
   \rowcolor{myrose}
   \tm{}-Inference-Only  & \textbf{60.24} & \textbf{69.07}\ci{1.47} & \textbf{63.78}\ci{4.69} \\
   \midrule
   Raw-Aligned & 3175.08 & 68.88\ci{1.48} & 55.80\ci{4.84} \\
   Summary-Aligned & 536.41 & 62.77\ci{1.54} & 45.09\ci{4.85} \\
   \rowcolor{myrose}
   \tm{}-Full  & \textbf{60.24} & \textbf{79.98}\ci{1.28} & \textbf{64.00}\ci{4.68} \\
   \bottomrule
   \end{tabular}
   \end{adjustbox}
   \label{tab:elix-ipo}
\end{table}

\noindent\textbf{Prompt-Dependent Summaries.}
Table~\ref{tab:elix-prompt-dependent} reports results for the prompt-dependent variant described in Section~\ref{sec:prompt-dependent-variant}, where the summary additionally conditions on the current prompt $x$. Conditioning on the prompt yields modest improvements over prompt-independent summaries, suggesting that most personalization signal is captured by user-level summaries derived from $s_i$ alone. This empirically supports our user-level scoping assumption: stable, prompt-independent preferences account for the majority of the personalization benefit.

\begin{table}[!htbp]
\caption{Performance comparisons on \texttt{ELIX} with the prompt-dependent preference summary. We report average additional input length (Avg. Len.), accuracy (Acc.), and Win Rate.}
\centering
\begin{adjustbox}{max width=0.55\linewidth}
\small
\begin{tabular}{lrrr}
\toprule
\multirow{2}{*}{Method} & \multicolumn{3}{c}{\texttt{ELIX} (Prompt-Dependent Summary)} \\
\cmidrule(lr){2-4}
& Avg. Len. $\downarrow$ & Acc. (\%) & Win Rate (\%) \\
\midrule
Base-Model & \multicolumn{1}{r}{--} & 50.00\ci{1.59} & 50.00\ci{4.88} \\
\midrule
Raw-Prompting & 3175.08 & 56.12\ci{1.58} & 49.31\ci{4.88} \\
Summary-Prompting & 543.66 & 54.18\ci{1.59} & 43.32\ci{4.83} \\
\rowcolor{myrose}
\tm{}-Inference-Only & \textbf{111.38} & \textbf{71.11}\ci{1.44} & \textbf{66.36}\ci{4.61} \\
\midrule
Raw-Aligned & 3175.08 & 73.94\ci{1.40} & 56.14\ci{4.84} \\
Summary-Aligned & 543.66 & 60.20\ci{1.56} & 43.06\ci{4.83} \\
\rowcolor{myrose}
\tm{}-Full & \textbf{111.38} & \textbf{81.65}\ci{1.23} & \textbf{65.37}\ci{4.64} \\
\bottomrule
\end{tabular}
\end{adjustbox}
\label{tab:elix-prompt-dependent}
\end{table}

\begin{table}[!ht]
    \caption{Ablation study on two-stage versus iterative co-training on \texttt{ELIX} (LLaMA-3.2-1B-Instruct) using Avg. Len., Acc., and Win Rate.}
    \centering
    \begin{adjustbox}{max width=\linewidth}
    \small
    \begin{tabular}{lrrrr}
    \toprule
    \multirow{2}{*}{Method} & \multirow{2}{*}{Training} & \multicolumn{3}{c}{\texttt{ELIX}} \\\cmidrule{3-5}
    & & Avg. Len. $\downarrow$ & Acc. (\%) & Win Rate (\%) \\
    \midrule
    Base-Model & -- & -- & 50.00\ci{1.59} & 50.00\ci{4.88} \\
    \midrule
    \multirow{2}{*}{\tm{}-Inference-Only } & Two-stage & \textbf{223.32} & \textbf{69.44}\ci{1.47} & \textbf{52.53}\ci{4.87} \\
    & Iterative & 2047.95 & 57.79\ci{1.57} & 16.98\ci{3.68} \\
    \midrule
    \multirow{2}{*}{\tm{}-Full } & Two-stage & \textbf{223.32} & \textbf{80.22}\ci{1.27} & \textbf{58.05}\ci{4.81} \\
    & Iterative & 2047.95 & 79.40\ci{1.29} & 54.63\ci{4.86} \\
    \bottomrule
    \end{tabular}
    \end{adjustbox}
    \label{tab:iterative}
    \end{table}

\noindent\textbf{Two-Stage vs.\ Iterative Co-Training.}
To empirically test the two-stage design choice discussed in Section~\ref{sec:two-stage}, we compare against a naive iterative variant that simply alternates between Stage~1 and Stage~2 over multiple rounds, with no constraint on how the two components co-adapt, using LLaMA-3.2-1B-Instruct for both models (Table~\ref{tab:iterative}). With fine-tuning, iterative training yields comparable Accuracy (79.40 vs.\ 80.22, with overlapping intervals) and Win Rate (54.63 vs.\ 58.05) but inflates summaries by roughly $9\times$ ($\sim$2{,}048 vs.\ $\sim$223 tokens), as co-adaptation removes the pressure to compress. The cost to transferability is severe: under a frozen base generator, iteratively trained summaries collapse to 16.98 Win Rate and 57.79 Accuracy---far below the two-stage results of 52.53 and 69.44---confirming that forgoing co-adaptation preserves the summary modularity underlying Tables~\ref{tab:elix-transfer}--\ref{tab:roleplay-transfer}.

\section{Conclusion}
\label{sec:conclusion}

We introduced \tm{}, a framework for user-level LLM personalization that separates the problem into preference inference and conditioned generation connected by a natural-language interface. A shared inference model, trained via reinforcement learning against a summary-augmented preference-optimization objective, distills heterogeneous user signals into concise natural-language preference summaries; a shared generator then conditions on these summaries to produce personalized responses. Our experiments across four benchmarks suggest that directly optimizing the preference-inference step---rather than treating it as heuristic preprocessing---is an effective lever for personalization: it produces compact summaries that reduce context overhead by a large margin while generally improving alignment with user preferences across the settings we tested. Because the interface is natural language, the learned summaries can be reused with frozen, off-the-shelf generators---including black-box commercial APIs---without retraining, and the optimization is complementary to existing handcrafted preference-extraction pipelines, providing further gains when applied as a refinement layer.

\noindent\textbf{Limitations and Future Work.}
The two-stage training design reflects a trade-off: optimizing the inference model against a frozen base generator encourages transferable summaries, but may compress away signal that a jointly fine-tuned generator could exploit. Iterative but constrained co-training that preserves transferability while retaining richer signal is a natural next step. More broadly, when user signals are especially sparse or heterogeneous, the extractable preference information may itself be limited; understanding when and why preference inference is structurally hindered under such conditions remains an open question. Finally, our framework assumes relatively stable user preferences; extending it to preferences that evolve over time would expand its applicability and connect to emerging work on continual adaptation in language models.

\subsubsection*{Broader Impact Statement}
This paper aims to advance the scientific understanding of user-level personalization in large language models through a modular framework for preference inference and conditioned generation. Potential societal impacts largely align with those typical of research on machine learning personalization: improved user experience, more adaptable model outputs, and broader accessibility for individuals with diverse needs. As with any personalization method, practical deployments should consider standard concerns such as privacy, transparency, and responsible use of user-provided signals. 

One consideration specific to our approach is that the learned preference summaries are natural-language text, which makes them human-readable. This is a double-edged property: on one hand, it enables transparency, since users or system operators can inspect and, if necessary, redact the summaries before they are transmitted to a generator; on the other hand, it means that preference-revealing content could be sent to third-party generator APIs in readable form. This concern applies broadly to any profile-conditioned system that calls external APIs, but our framework makes it more salient because the summaries are interpretable text rather than opaque embeddings. In practice, several mitigations apply: in our experiments, the summaries we inspected largely captured stylistic and topical preferences rather than personally identifiable information, though nothing in the objective enforces this, and user signals may themselves contain demographic attributes; users can review and edit the summaries precisely because they are readable; and standard privacy-preserving techniques such as differential privacy or on-device inference can be layered on top.

% % Add author contributions here once the submission is accepted and deanonymized.

% % Add acknowledgments here once the submission is accepted and deanonymized.

\bibliography{main}
\bibliographystyle{tmlr}

\appendix

\section*{Appendix}

\section{Training and Inference Algorithms}\label{appendix:alg}

Algorithms~\ref{alg:popi-training} and~\ref{alg:popi-inference} summarize the two-stage training procedure and the inference procedure of \tm{}, respectively, as described in Sections~\ref{sec:dpo} and~\ref{sec:two-stage}.

\begin{algorithm}[!ht]
    \caption{Training Procedure of \tm{}}
    \label{alg:popi-training}
    \begin{algorithmic}[1]
    \INPUT User signals $\{s_i\}_{i=1}^N$, preference-labeled datasets $\{\mathcal{D}_i\}_{i=1}^N$, reference inference model $\pi_{\phi_{\text{ref}}}$, reference generator $\pi_{\theta_{\text{ref}}}$
    \OUTPUT Trained inference model $\pi_\phi$, (optional) trained generator $\pi_\theta$
    
    \STATE Initialize $\pi_\phi \leftarrow \pi_{\phi_{\text{ref}}}$
    \STATE Initialize $\pi_\theta \leftarrow \pi_{\theta_{\text{ref}}}$
    
    \vspace{0.25em}
    \STATE \textbf{Stage 1: Optimizing $\pi_\phi$}
    
    \FOR{each training iteration}
        \STATE Sample user $i$ and example $(x,y_c,y_r)\sim\mathcal{D}_i$
        \STATE Sample a group of preference summaries $z_i \sim \pi_\phi(\cdot \mid s_i)$
        \FOR{each sampled $z_i$}
            \STATE Compute the reward as the negative summary-augmented objective $-\ell_i$ using Equation~\ref{eq:sadpo}
        \ENDFOR
        \STATE Update $\pi_\phi$ with \textbf{GRPO}, with regularization $
        \alpha\,\mathrm{KL}\!\left(\pi_\phi(\cdot\mid s_i)\,\|\,\pi_{\phi_\text{ref}}(\cdot\mid s_i)\right)$
    \ENDFOR
    
    \vspace{0.25em}
    \STATE \textbf{Stage 2 (Optional): Optimizing $\pi_\theta$}
    \FOR{each training iteration}
        \STATE Sample user $i$ and example $(x,y_c,y_r)\sim\mathcal{D}_i$
        \STATE Sample preference summary $z_i \sim \pi_\phi(\cdot \mid s_i)$
        \STATE Compute the summary-augmented objective $\ell_i$ using Equation~\ref{eq:sadpo}
        \STATE Update $\pi_\theta$ by minimizing $\ell_i$ via standard backpropagation
    \ENDFOR
    \end{algorithmic}
    \end{algorithm}
    
    \begin{algorithm}[!ht]
    \caption{Inference Procedure of \tm{}}
    \label{alg:popi-inference}
    \begin{algorithmic}[1]
    \INPUT User signals $s_i$, user prompt(s) $\{x\}$, trained inference model $\pi_\phi$, trained, frozen, or black-box generator $\pi_\theta$
    \OUTPUT Personalized response(s) $\{y\}$
    
    \STATE Generate a preference summary $z_i \sim \pi_\phi(\cdot \mid s_i)$
    \FOR{each user prompt $x$}
        \STATE Generate a personalized response $y \sim \pi_\theta(\cdot \mid x, z_i)$
    \ENDFOR
    \STATE \textbf{return} $\{y\}$
    \end{algorithmic}
    \end{algorithm}

\section{Derivation of the Decomposition in Equation~\ref{eq:bound}}
\label{appendix:bound}

We derive the information-theoretic decomposition of the summary-augmented DPO objective $\mathcal{L}$. Following the standard treatment in~\citet{rafailov2023direct}, we begin by deriving an implicit reward model parameterized by the generation model $\pi_\theta$:
\begin{equation}
\begin{aligned}
    r_{\pi_\theta}(x,y,z_i) = \beta \log \frac{\pi_{\theta}(y \mid x, z_i)}{\pi_{\theta_\text{ref}}(y \mid x)} + \beta \log Z(x,z_i), \\
    Z(x,z_i) = \sum_y \pi_{\theta_\text{ref}}(y \mid x) \exp(\frac{1}{\beta}r_{\pi_\theta}(x,y,z_i)).
\end{aligned}
\end{equation}

Under the Bradley--Terry (BT) model of preferences, the conditional probability that $y_c$ is preferred over $y_r$, given prompt $x$ and summary $z_i$, is expressed as:
\begin{equation}
P_{\pi_\theta}(y_c \succ y_r \mid x,z_i) = \sigma \Big( r_{\pi_\theta}(x,y_c,z_i) - r_{\pi_\theta}(x,y_r,z_i) \Big).
\end{equation}

Substituting this into the $\mathcal{L}$ formulation, we can write it as:
\begin{equation}
    \mathcal{L} = \frac{1}{N} \sum_{i=1}^N \ \mathbb{E}_{(x, y_c, y_r) \sim \mathcal{D}_i,\ z_i \sim \pi_{\phi}(\cdot \mid s_i)} \Big[ -\log P_{\pi_\theta}(y_c \succ y_r \mid x,z_i) \Big].
\end{equation}

We then decompose this objective to reveal its information-theoretic structure:
\begin{equation}
\begin{aligned}
\mathcal{L}
= & \ \frac{1}{N} \sum_{i=1}^N \ \mathbb{E}_{(x, y_c, y_r), z_i}\Bigg[ \log \frac{P(y_c \succ y_r \mid x,z_i)}{P_{\pi_\theta}(y_c \succ y_r \mid x,z_i)} \\
& \quad + \log \frac{1}{P(y_c \succ y_r \mid x)} - \log \frac{P(y_c \succ y_r \mid x,z_i)}{P(y_c \succ y_r \mid x)} \Bigg] \\
= & \ \text{KL}\!\left(P(y_c \succ y_r \mid x, z_i) \,\|\, P_{\pi_\theta}(y_c \succ y_r \mid x,z_i)\right) \\
& \quad + H(P(y_c \succ y_r \mid x)) - I(y_c \succ y_r ; z_i \mid x),
\end{aligned}
\end{equation}
which is the decomposition stated in Equation~\ref{eq:bound}. The entropy term $H(P(y_c \succ y_r \mid x))$ is independent of both models' parameters and therefore acts as a constant throughout training, so optimization of $\mathcal{L}$ reduces to jointly minimizing the KL term (generator approximation error) and maximizing the conditional mutual information (summary informativeness), as discussed in the main text.

\section{Stability of RL Training}\label{appendix:rl}
\begin{figure}[!ht]
    \centering
    \includegraphics[width=0.5\linewidth]{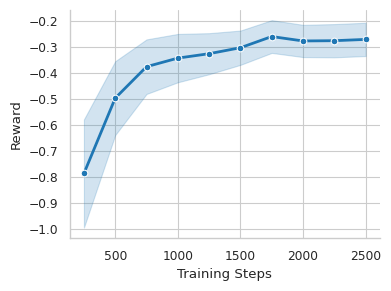}
    \caption{Mean reward (solid line) and standard deviation (shaded area) during GRPO training on the \texttt{Review} dataset.}
    \label{fig:reward}
\end{figure}

We analyze the stability of RL by tracking the mean and standard deviation of the reward during training on the \texttt{Review} dataset. As shown in Figure~\ref{fig:reward}, the average reward increases steadily, while the variance decreases and stabilizes over time. This behavior indicates stable and well-behaved optimization of the preference inference model under our training setup.

\section{Ablation: Effect of the Scaling Parameter}\label{appendix:beta}

Table~\ref{tab:elix-beta} examines sensitivity to the hyperparameter $\beta \in \{0.1, 0.05, 0.01\}$, which scales the implicit reward model. \tm{} performs consistently across this range, with $\beta = 0.01$ yielding the best results.

\begin{table}[!ht]
\caption{Ablation study on the hyperparameter $\beta$ on \texttt{ELIX} using Avg. Len., Acc., and Win Rate.}
\centering
\begin{adjustbox}{max width=\linewidth}
\small
\begin{tabular}{lrrrr}
\toprule
\multirow{2}{*}{Method} & \multirow{2}{*}{$\beta$} & \multicolumn{3}{c}{\texttt{ELIX}} \\\cmidrule{3-5}
& & Avg. Len. $\downarrow$ & Acc. (\%) & Win Rate (\%) \\
\midrule
Base-Model & -- & -- & 50.00\ci{1.59} & 50.00\ci{4.88} \\
\midrule
\multirow{3}{*}{\tm{}-Inference-Only } & 0.10 & 118.80 & 68.43\ci{1.48} & 59.92\ci{4.78} \\
& 0.05 & 145.92 & 70.42\ci{1.45} & 62.97\ci{4.71} \\
& 0.01 & \textbf{52.52} & \textbf{71.48}\ci{1.44} & \textbf{63.84}\ci{4.69} \\
\midrule
\multirow{3}{*}{\tm{}-Full } & 0.10 & 118.80 & 76.54\ci{1.35} & 53.91\ci{4.86} \\
& 0.05 & 145.92 & 79.69\ci{1.28} & 59.05\ci{4.80} \\
& 0.01 & \textbf{52.52} & \textbf{80.14}\ci{1.27} & \textbf{63.97}\ci{4.68} \\
\bottomrule
\end{tabular}
\end{adjustbox}
\label{tab:elix-beta}
\end{table}

\section{Ablation: Effect of the KL Regularization Weight}\label{appendix:alpha}

Table~\ref{tab:elix-alpha} examines sensitivity to the coupling constant $c$ in $\alpha = c \cdot \beta$, varied over a 5$\times$ range with $\beta$ and all other hyperparameters held fixed. Quality metrics are stable across this range: Accuracy varies by less than three points, and Win Rate is comparable for $c = 0.002$ and $c = 0.005$, with overlapping intervals. Summary length is more sensitive to $c$, and the value used throughout the paper ($c = 0.002$) yields both the shortest summaries and the highest Win Rate.

\begin{table}[!ht]
\caption{Ablation study on the coupling constant $c$ in $\alpha = c \cdot \beta$ on \texttt{ELIX} using Avg. Len., Acc., and Win Rate.}
\centering
\begin{adjustbox}{max width=\linewidth}
\small
\begin{tabular}{lrrrr}
\toprule
\multirow{2}{*}{Method} & \multirow{2}{*}{$c$} & \multicolumn{3}{c}{\texttt{ELIX}} \\\cmidrule{3-5}
& & Avg. Len. $\downarrow$ & Acc. (\%) & Win Rate (\%) \\
\midrule
Base-Model & -- & -- & 50.00\ci{1.59} & 50.00\ci{4.88} \\
\midrule
\multirow{3}{*}{\tm{}-Inference-Only } & 0.001 & 197.80 & \textbf{71.74}\ci{1.43} & 56.81\ci{4.83} \\
& 0.002 & \textbf{52.52} & 71.48\ci{1.44} & \textbf{63.84}\ci{4.69} \\
& 0.005 & 132.03 & 69.09\ci{1.47} & 63.61\ci{4.69} \\
\bottomrule
\end{tabular}
\end{adjustbox}
\label{tab:elix-alpha}
\end{table}

\section{Ablation: Robustness to Prompting Templates}\label{appendix:template}

To assess whether our results depend on a particular prompting template, we re-ran the generator-transferability experiments on \texttt{ELIX} using a different set of templates (Figure~\ref{fig:prompts2}). Table~\ref{tab:elix-transfer-new-template} reports the results: although individual models exhibit some variance across templates, the averaged performance remains consistent, indicating that the gains from optimized preference summaries are not driven by template-specific effects.

\renewcommand{\heatmap}[1]{%
  \begingroup
  \edef\temp{#1}%
  \expandafter\heatmapaux\temp\relax
  \endgroup
}

\def\heatmapaux#1(#2\relax{%
  \cellcolor{myrose!\fpeval{(#1 - 19.02)*100/64.93}!myblue}%
  #1(#2%
}

\begin{table}[!ht]
\caption{Comparison of win rates on \texttt{ELIX} using an alternative prompting template set, across off-the-shelf LLMs. The win rates are computed relative to each LLM's own Base-Model. Cells are color-coded, where darker blue indicates lower values, and lighter red indicates higher values. Numbers in parentheses indicate changes relative to the original templates. Confidence intervals are omitted here for space, as each cell already reports a paired change; the per-generator sample size matches Table~\ref{tab:elix-transfer} ($N=200$).}
\centering
\begin{adjustbox}{max width=\linewidth}
\small
\begin{tabular}{lccccccccc}
\toprule
\multirow{2}{*}{Method} & \multicolumn{8}{c}{Win Rate (\%) on \texttt{ELIX}} & \multirow{2}{*}{Avg.} \\
\cmidrule{2-9}
& Mistral-S & Mistral-L & DeepSeek-R1 & Llama-1b & Llama-11b & Llama-90b & Claude-4 & GPT-4o-mini & \\
\midrule
Base-Model
& \heatmap{50.00(+0.00)} & \heatmap{50.00(+0.00)} & \heatmap{50.00(+0.00)}
& \heatmap{50.00(+0.00)} & \heatmap{50.00(+0.00)} & \heatmap{50.00(+0.00)}
& \heatmap{50.00(+0.00)} & \heatmap{50.00(+0.00)} & \heatmap{50.00(+0.00)} \\
Raw-Prompting
& \heatmap{58.23(-9.04)} & \heatmap{74.00(+1.84)} & \textbf{\heatmap{77.28(-2.10)}}
& \heatmap{31.38(+4.22)} & \heatmap{58.70(+2.65)} & \textbf{\heatmap{57.98(-1.05)}}
& \textbf{\heatmap{75.47(+3.09)}} & \heatmap{69.57(-1.86)} & \heatmap{62.83(-0.28)} \\
Summary-Prompting
& \heatmap{56.17(-4.96)} & \heatmap{65.50(-1.65)} & \heatmap{69.22(-5.31)}
& \heatmap{19.02(+0.27)} & \heatmap{49.06(+0.42)} & \heatmap{50.00(+1.67)}
& \heatmap{59.74(-0.22)} & \heatmap{65.65(-2.20)} & \heatmap{54.30(-1.50)} \\
\rowcolor{myrose}
POPI-Inference-Only
& \textbf{\heatmap{75.98(-5.25)}} & \textbf{\heatmap{83.95(+2.45)}} & \heatmap{70.32(-0.56)}
& \textbf{\heatmap{67.93(+5.28)}} & \textbf{\heatmap{66.10(+2.57)}} & \heatmap{57.88(-1.74)}
& \heatmap{70.63(+0.41)} & \textbf{\heatmap{82.76(+6.23)}} & \textbf{\heatmap{71.94(+1.17)}}\\
\bottomrule
\end{tabular}
\end{adjustbox}
\label{tab:elix-transfer-new-template}
\end{table}

\newpage

\section{Qualitative Illustration: Summary Before vs.\ After Optimization}\label{appendix:summary-case}

\begin{figure}[!ht]
    \centering
    \includegraphics[width=\linewidth]{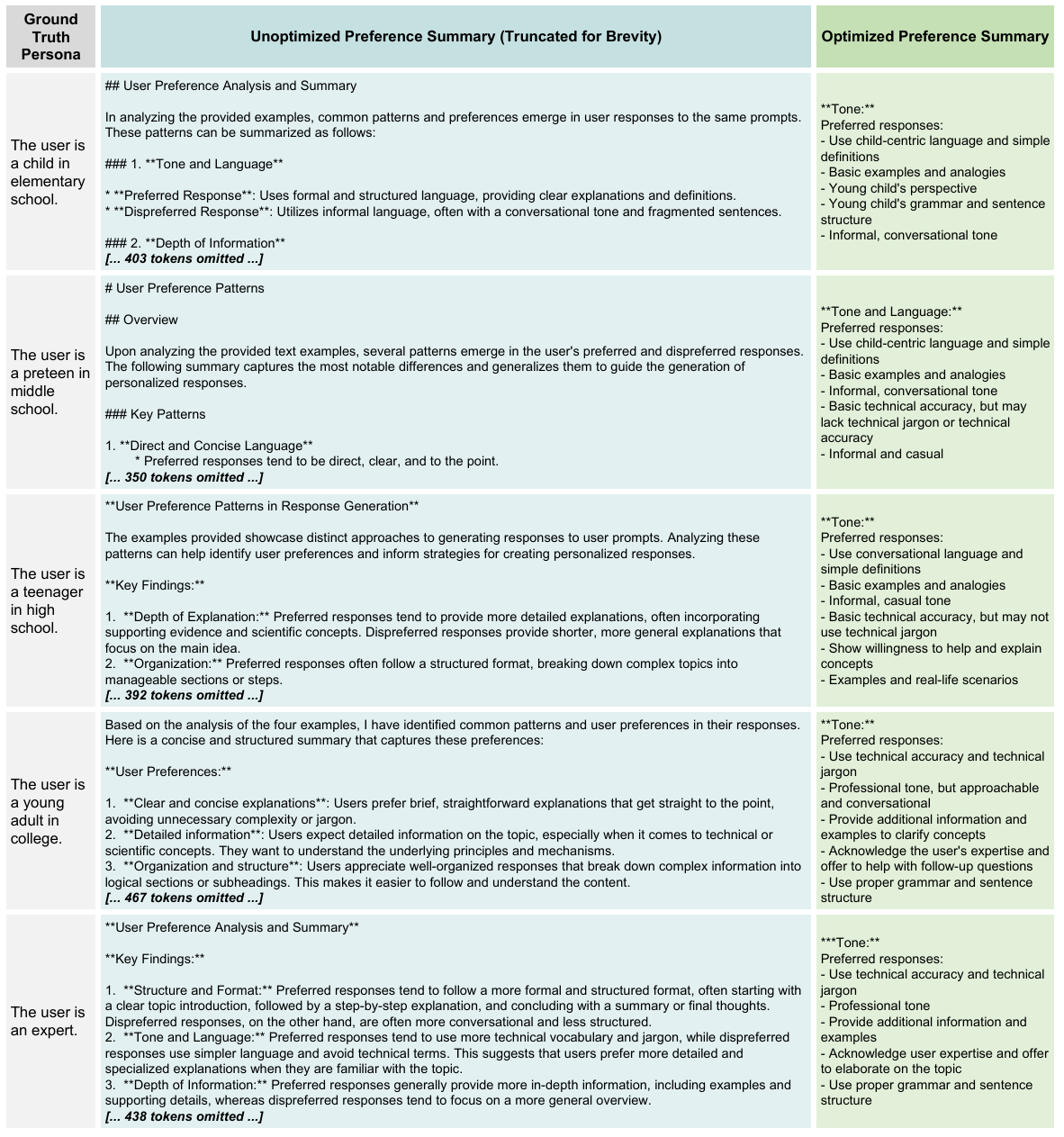}
    \caption{Comparison of preference summaries before and after optimization on the \texttt{ELIX} dataset (same thermostat example as Figure~\ref{fig:case}). Unoptimized summaries are truncated for space; token counts are annotated. The unoptimized inference model produces verbose, structurally generic summaries across all personas, while the optimized model produces compact, persona-specific summaries that capture distinct preference profiles.}
    \label{fig:summary-case}
\end{figure}

\newpage

\section{Qualitative Illustration: Summary and Personalized Generation}\label{appendix:case}

\begin{figure}[!ht]
    \centering
    \includegraphics[width=\linewidth]{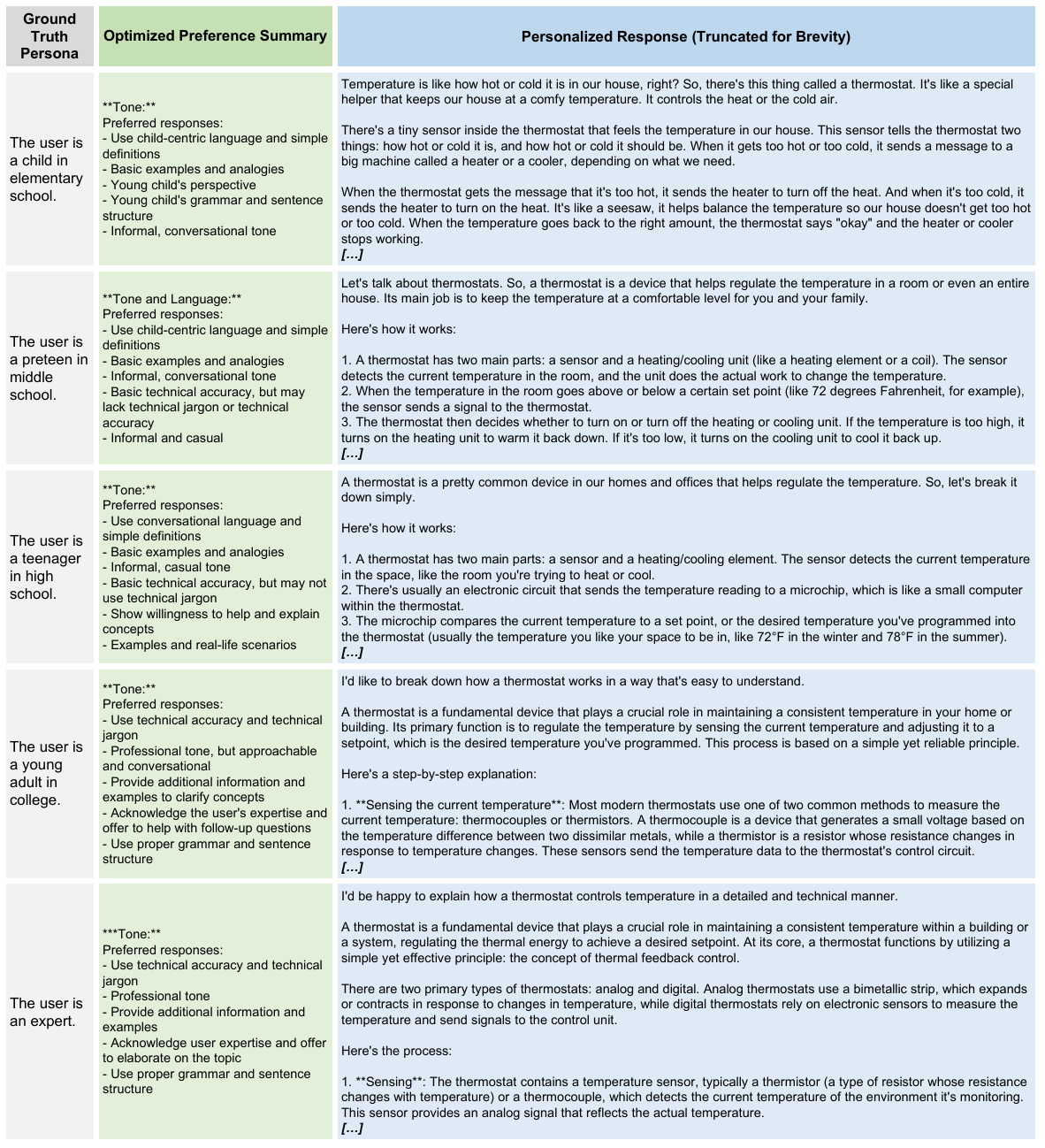}
    \caption{Qualitative case study on the ELIX dataset. Each row corresponds to one of the five ground-truth personas in ELIX, all responding to the same user prompt: ``How does a thermostat control temperature?'' The ground-truth personas are not accessible during training. For each persona, we show an example of the optimized preference summary inferred from user signals, and the corresponding personalized response generated (truncated for brevity). The summaries capture stylistic and content-level preferences at different expertise levels, while the generated responses demonstrate how conditioning on these summaries yields appropriately tailored explanations.}
    \label{fig:case}
\end{figure}

\newpage

\section{Qualitative Illustration: \texttt{Review} and \texttt{Roleplay}}\label{appendix:case-cross}

\begin{figure}[!ht]
    \centering
    \includegraphics[width=\linewidth]{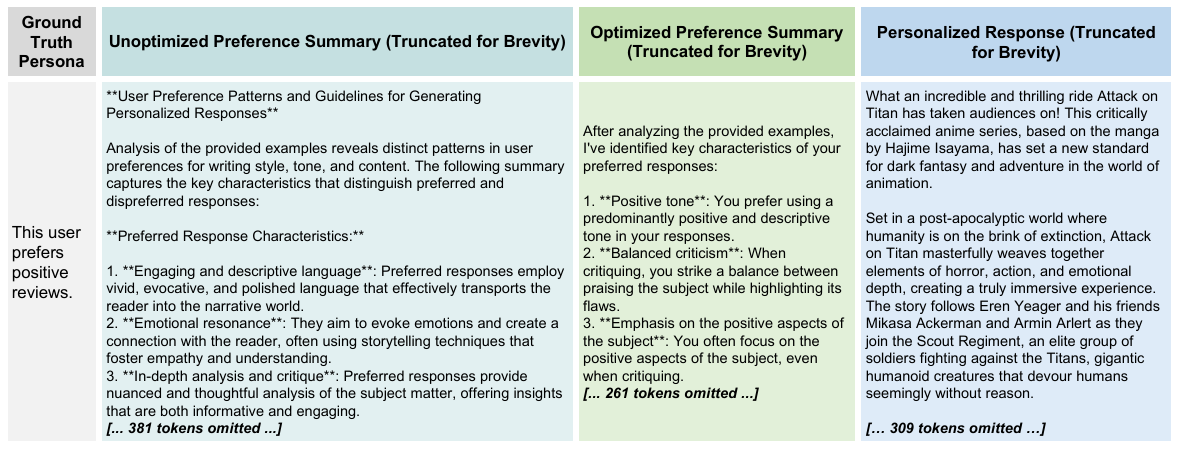}
    \caption{Qualitative case study on the \texttt{Review} dataset, for the user prompt: ``Write a review for the Animation TV show, Attack on Titan.'' The ground-truth persona is not accessible during training. We show the preference summary before and after optimization, together with the personalized response generated from the optimized summary (truncated for brevity where indicated, with omitted token counts annotated). The unoptimized summary describes generic qualities of a well-written review, while the optimized summary recovers the user's preference for positive reviews.}
    \label{fig:review-case}
\end{figure}

\begin{figure}[!ht]
    \centering
    \includegraphics[width=\linewidth]{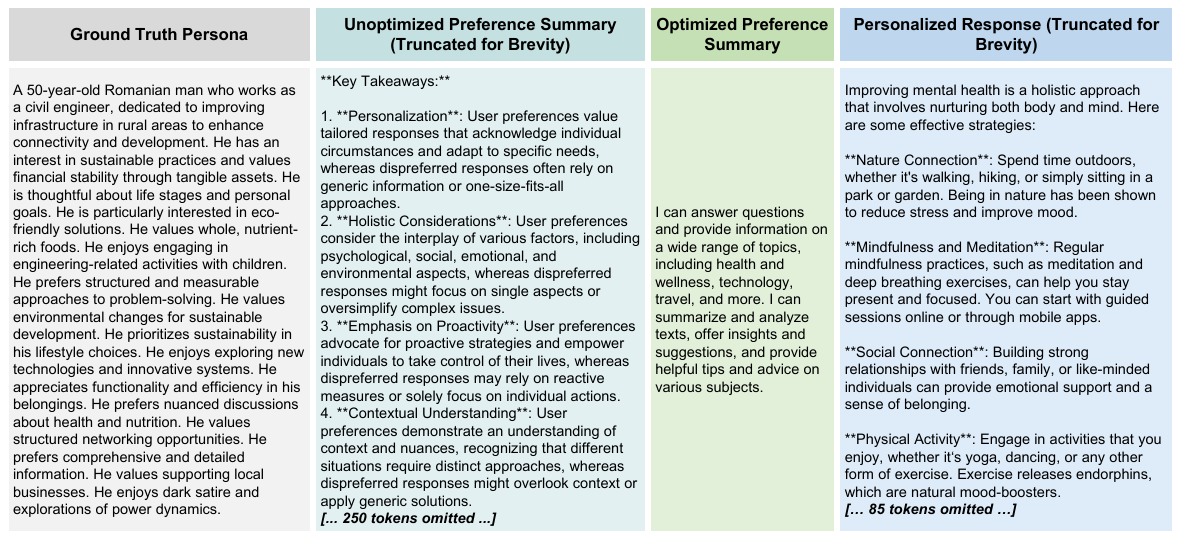}
    \caption{Qualitative case study on the \texttt{Roleplay} dataset, for the user prompt: ``What are some strategies to improve my mental health?'' The ground-truth persona is not accessible during training. We show the preference summary before and after optimization, together with the personalized response generated from the optimized summary (truncated for brevity where indicated, with omitted token counts annotated). The optimized summary distills the user signals into a few sentences about the user's general topical interests, reflecting the limited preference information carried by \texttt{Roleplay}'s user signals. We note that it is phrased in the first person, as an assistant capability statement rather than a description of the user; this reflects the reference model's instruction-tuned prior, toward which the KL-regularized objective pulls the summary when the reward signal is weak.}
    \label{fig:roleplay-case}
\end{figure}

\newpage

\section{Preference Inference and Personalized Generation Prompting Templates}\label{appendix:prompts}
\begin{figure}[!ht]
    \centering
    \includegraphics[width=\linewidth]{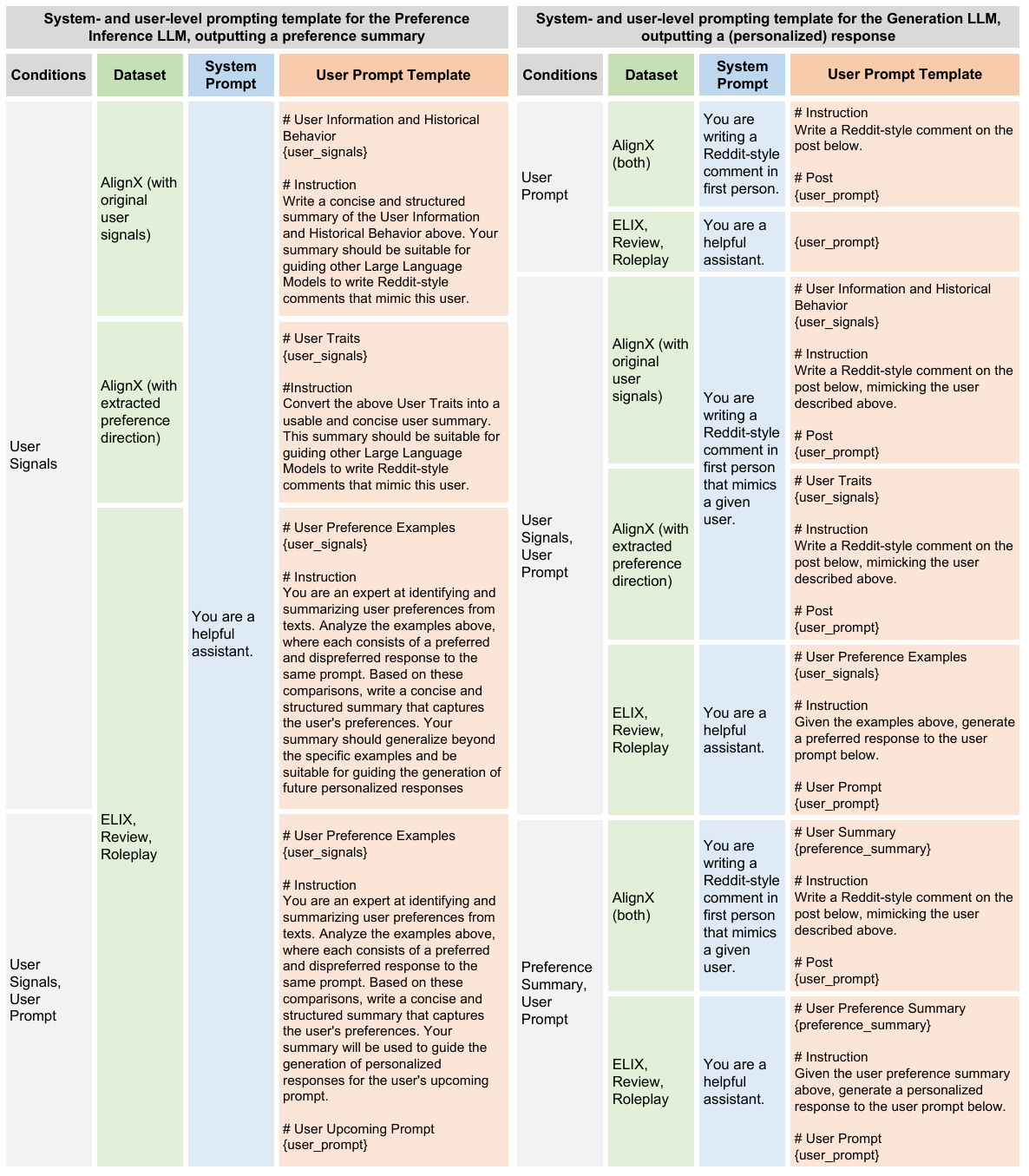}
    \caption{Left: prompting templates for the preference inference LLM, which transforms raw user signals (or user signals plus prompt) into natural language summaries. Right: prompting templates for the generation LLM, which integrate user prompts, raw signals, or preference summaries to produce personalized responses. These templates render conditioning variables as natural language.}
    \label{fig:prompts}
\end{figure}

\newpage

\section{Alternative Personalized Generation Prompting Templates for Table~\ref{tab:elix-transfer-new-template}}\label{appendix:prompts2}

\vspace{-0.1in}

\begin{figure}[!htbp]
    \centering
    \includegraphics[width=\linewidth]{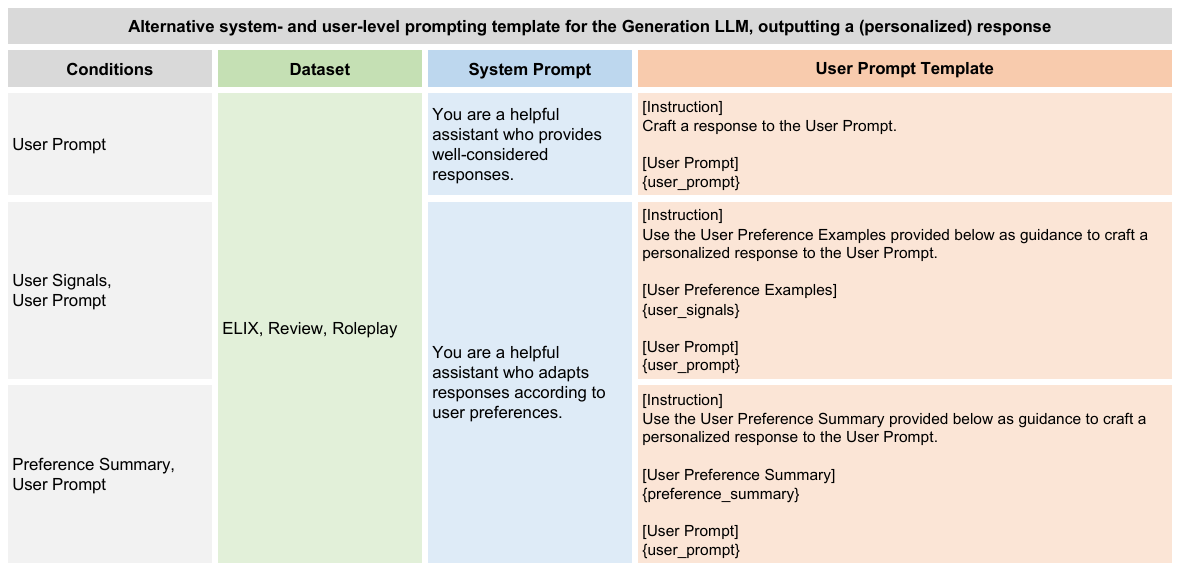}
    \caption{Alternative prompting templates for the generation LLM, used in the robustness ablation study (Section~\ref{sec:ablation}, Table~\ref{tab:elix-transfer-new-template}). These templates differ from the primary templates shown in Figure~\ref{fig:prompts} in structure, wording, and length, while preserving the same conditioning variables.}
    \label{fig:prompts2}
\end{figure}

\vspace{-0.1in}

\section{LLM-as-a-Judge Prompting Template}
\label{appendix:judge-prompt}
\vspace{-0.1in}
\begin{figure}[!ht]
    \centering
    \includegraphics[width=\linewidth]{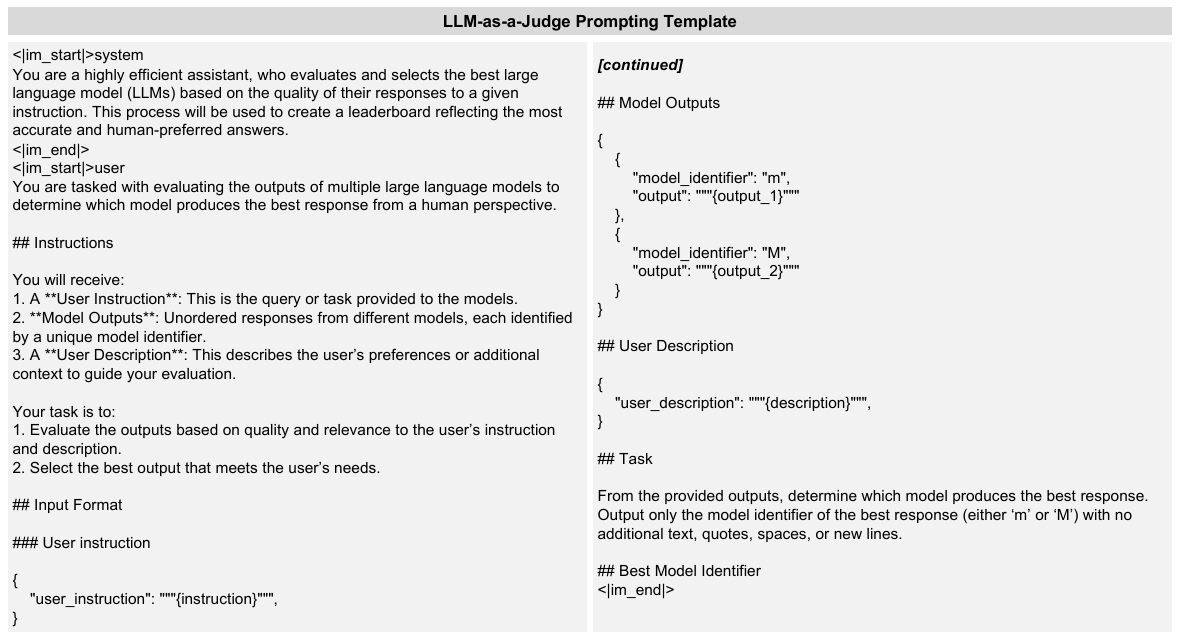}
    \caption{The template used when querying GPT-4o to serve as a judge. The template ensures that the model considers both the user's prompt and the ground-truth user description when comparing candidate outputs.}
    \label{fig:llmaaj}
\end{figure}

\end{document}